\documentclass[10pt,twocolumn,letterpaper]{article}

\usepackage[pagenumbers]{cvpr} 

\usepackage{graphicx}
\usepackage{amsmath}

\usepackage[utf8]{inputenc} 
\usepackage[T1]{fontenc}    
\usepackage[pagebackref,breaklinks,colorlinks]{hyperref}
\usepackage{url}            
\usepackage{booktabs}       
\usepackage{amsfonts}       
\usepackage{nicefrac}       
\usepackage{microtype}      
\usepackage{xcolor}         
\usepackage{wrapfig}
\usepackage[normalem]{ulem}
\usepackage{amsmath}
\usepackage{amssymb}
\usepackage{multirow}
\usepackage{paralist}
\usepackage{float}

\newtheorem{prop}{Proposition}

\usepackage[permil]{overpic}
\usepackage[linesnumbered,ruled]{algorithm2e}

\def\ie{\emph{i.e.}}
\def\eg{\emph{e.g.}}

\def\etc{{\em etc.}}

\usepackage[capitalize]{cleveref}
\crefname{section}{Sec.}{Secs.}
\Crefname{section}{Section}{Sections}
\Crefname{table}{Table}{Tables}
\crefname{table}{Tab.}{Tabs.}

\usepackage{pifont}

\newcommand{\methodname}{SSP}

\newcommand{\cmark}{\ding{51}}
\newcommand{\bluecmark}{{\color{blue}\cmark}}

\newcommand{\xmark}{\ding{55}}
\newcommand{\redxmark}{{\color{red}\xmark}}

\def\cvprPaperID{XXXX} 
\def\confName{CVPR}
\def\confYear{2023}

\title{Semi-signed prioritized neural fitting for surface reconstruction\\ from unoriented point clouds}

\def \cuhk{$^{1}$}
\def \tencentAI{$^{2}$}
\def \whu{$^{3}$}

\author{%
  Runsong Zhu$^{1}$\footnotemark[1]\quad
  Di Kang\tencentAI\quad
  Ka-Hei Hui\cuhk\quad
  Yue Qian\tencentAI\quad
  Xuefei Zhe\tencentAI\\
  Zhen Dong\whu\footnotemark[2]\quad
  Linchao Bao\tencentAI\quad
  Pheng-Ann Heng\cuhk\quad
  Chi-Wing Fu\cuhk\quad
  \\
  \cuhk The Chinese University of Hong Kong \quad 
  \tencentAI Tencent AI Lab\quad
  \whu Wuhan University
  \\
}

\begin{document}
\twocolumn[{
\maketitle
\begin{figure}[H]
\hsize=\textwidth
\centering
\includegraphics[width=\textwidth]{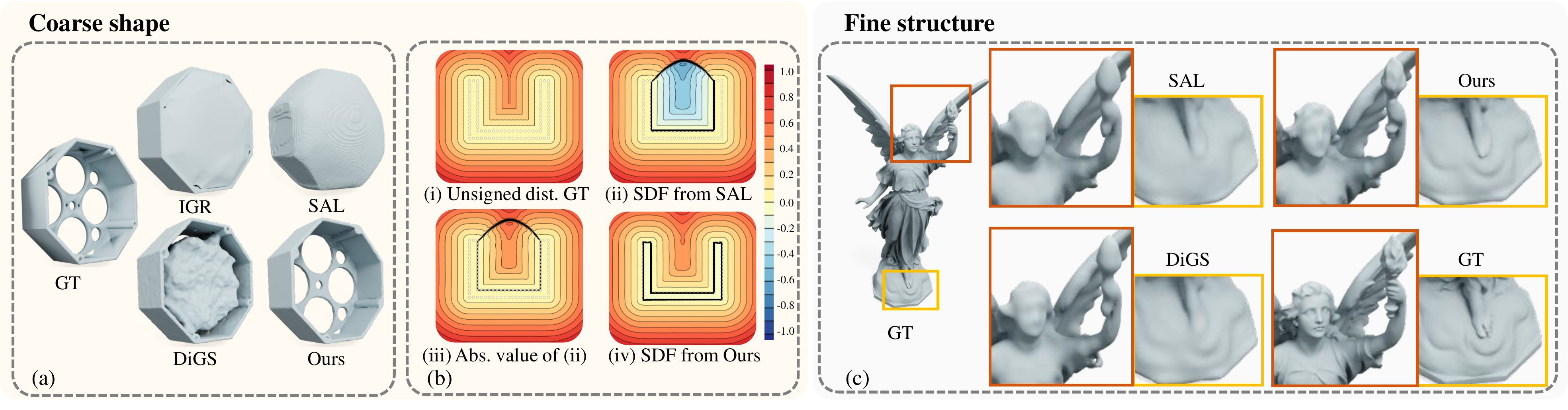}
\caption{
\textbf{(a)} 
Existing methods occasionally suffer from finding the coarse shape;
\textbf{(b)}  
A 2D example that illustrates this ambiguous issue (using only the ambiguous unsigned supervision).
Given the GT unsigned distance field in (b-i) as constraints, 
SAL~\cite{atzmon2020sal} yields an SDF (b-ii) whose absolute-valued distance field (b-iii) is quite close to its GT, but with undesired surfaces (b-ii).
Ours (b-iv) does not suffer from this issue due to the help of signed supervision in the ``outside'' region; and
\textbf{(c)} Comparisons on a challenging case with other baselines~\cite{atzmon2020sal,ben2022digs} that utilize a random sampling strategy demonstrate that our proposed loss-based region sampling is good at reconstructing fine structures.
}
\label{fig:teaser}
\end{figure}
}]

\renewcommand{\thefootnote}{\fnsymbol{footnote}}
\footnotetext[1]{Work partially done during an internship at Tencent AI Lab.}
\footnotetext[2]{Corresponding author.}

\begin{abstract}
Reconstructing 3D geometry from \emph{unoriented} point clouds can benefit many downstream tasks.
Recent shape modeling methods mostly adopt implicit neural representation to fit a signed distance field (SDF) and optimize the network by \emph{unsigned} supervision.
However, these methods occasionally have difficulty in finding the coarse shape for complicated objects, especially suffering from the ``ghost'' surfaces (\ie, fake surfaces that should not exist).
To guide the network quickly fit the coarse shape, we propose to utilize the signed supervision in regions that are obviously outside the object and can be easily determined, resulting in our semi-signed supervision.
To better recover high-fidelity details, a novel importance sampling based on tracked region losses and a progressive positional encoding (PE) prioritize the optimization towards underfitting and complicated regions.
Specifically, we voxelize and partition the object space into \emph{sign-known} and \emph{sign-uncertain} regions, in which different supervisions are applied.
Besides, we adaptively adjust the sampling rate of each voxel according to the tracked reconstruction loss, so that the network can focus more on the complicated under-fitting regions.
To this end, we propose our semi-signed prioritized (SSP) neural fitting, and conduct extensive experiments to demonstrate that {\methodname} achieves state-of-the-art performance on multiple datasets including the ABC subset and various challenging data.
The code will be released upon the publication.
\end{abstract}



\vspace{-3mm}
\section{Introduction}
\label{sec:intro}

Surface reconstruction from \emph{unoriented} point clouds is a long-standing fundamental task for many downstream applications in computer vision, computer graphics, and AR/VR.
However, due to the unstructured data format of point clouds, it remains challenging to reconstruct accurate surfaces for complicated topology-agnostic objects.

Among various approaches, implicit methods have gained increasing interest as they can reconstruct smooth and high-fidelity surfaces.
Traditional implicit methods reconstruct surfaces by calculating global (\eg, RBF~\cite{carr2001reconstruction}, SPSR~\cite{kazhdan2013screened}) or local (\eg, IMLS~\cite{shen2004interpolating}) implicit functions.
However, these methods suffer from cumbersome pre-processing (e.g., denoising, upsampling, normal estimation, normal orientation, etc.), among which accurately oriented normals contribute a lot to high-quality reconstructions.
Unfortunately, it is notoriously hard to compute orientation information from point clouds~\cite{berger2017survey}, thus limiting these methods' applicability.

\par
Recently, significant progress~\cite{peng2021shape,ma2021pull,gropp2020reg,atzmon2020sal,atzmon2020sald,atzmon2019controlling,ben2022digs} has been made in directly optimizing an implicit function (\eg, SDF) from unoriented point clouds.
For example, \cite{ma2021pull,gropp2020reg,atzmon2020sal,atzmon2020sald,atzmon2019controlling,ben2022digs} explore unsigned supervision (due to lack of GT except the points) to optimize the neural network and demonstrate promising reconstruction results.
If the optimization process goes well, the neural network will fit a coarse shape (\eg, an overly-smoothed hull of the object) at the early stage and then recover the fine structures.
Although substantial improvements have been achieved by existing methods~\cite{atzmon2020sal,atzmon2020sald,gropp2020reg,ben2022digs},
there remain several challenges that prevent them from producing high-quality reconstructions.

%
%
First, most existing methods occasionally have difficulty in finding coarse shapes for complicated objects,
leading to the production of \emph{``ghost''} surfaces (\ie, fake surfaces that should not exist) in undesired locations and large errors (see Fig.~\ref{fig:teaser}(a)).
This implies that the existing unsigned supervision may not be able to provide sufficient guidance, so as the network occasionally gets stuck at bad local minimums and generates ghost structures.
A 2D synthetic example is shown in Fig.~\ref{fig:teaser}(b). 
It satisfies the unsigned distance supervision~\cite{atzmon2020sal} very well but a fake surface still appears, which means using only unsigned distance supervision is insufficient for certain cases.
%
The second issue we observe is that most existing methods tend to reconstruct over-smoothing surfaces in complicated regions (\eg, containing details in different levels) and ignore some fine structures (see,~\eg, Fig.~\ref{fig:teaser}(c)).
%

In this paper, we make a step toward overcoming those two problems.
For the ``ghost-surface'' problem, we propose a simple yet effective solution by introducing an extra coarse signed supervision.
The insight is that signed supervision is more informative and we can apply signed supervision to regions that are apparently ``outside'' the target object.
Specifically, we propose a novel \emph{semi-signed} fitting module, which simultaneously provides coarse signed supervisions and unsigned supervisions for different regions determined by our automatic \emph{space partitioning} algorithm. 
With the additional signed guidance, the network can quickly fit a coarse shape to the given point cloud in the early stage, 
thus having higher chance in avoiding potential sub-optimal local minimums (within the computation budget). 
For the lack of details, we propose a new importance sampling strategy to increase the optimization efficiency and utilize progressive positional encoding to better recover the details.
Specifically, we design a new loss-based region sampling (LRS) strategy that tracks the losses in the full 3D space and adaptively increases the sampling density of regions with larger losses.
As the semi-signed fitting module can effectively avoid bad local minimums,
LRS helps the network focus on the details without suffering from severe ghost surface issues.
Considering that MLP may have difficulty in fitting high-frequency signals~\cite{tancik2020fourier},
we explore progressive positional encoding (PE) to further improve the reconstruction details.
%
%
Overall, we propose semi-signed prioritized (SSP) neural fitting for more stable and accurate surface reconstruction from raw point clouds.
To evaluate its effectiveness, we conduct extensive experiments.
Our SSP achieves state-of-the-art accuracy on multiple datasets,
including the ABC subset~\cite{erler2020points2surf} and
various challenging data (\ie, objects with complicated structures~\cite{zhou2016thingi10k}, objects with varying sampling density~\cite{guerrero2018pcpnet}, and objects with sampling noise~\cite{guerrero2018pcpnet}).
We also conduct several experiments to show that our proposed signed supervision can be adopted in existing methods (\eg, IGR~\cite{gropp2020reg} and DiGS~\cite{ben2022digs}) to boost their accuracy.
%

%
%
Our contributions are summarized as follows:
\begin{compactitem}
\item We propose a new semi-signed fitting module that provides additional signed supervision, which significantly alleviates the difficulty in finding coarse shapes for complicated objects.
\item We introduce a loss-based per-region sampling and progressive PE, resulting in accurate surfaces with more details while generating fewer artifacts.
\item We propose semi-signed prioritized  (SSP) neural fitting, achieving improved performances on multiple datasets, especially with significant reduction on CD-$L_{1}$ over existing neural fitting methods (\eg, $20\%$ improvement on ABC subset~\cite{erler2020points2surf} upon existing SOTA DiGS~\cite{ben2022digs}).
\end{compactitem}


\section{Related Work}

Reconstructing surfaces from unoriented point clouds is a long-standing problem.
Here, we review the existing methods from traditional methods to learning-based and optimization-based neural methods.

\vspace{-3mm}
\paragraph{Traditional methods.}
Early methods address the reconstruction task based on either handcrafted heuristics or numerical optimizations.
In particular, some adopt heuristic guidance to progressively build and refine the reconstructed surface, such as 
growing triangulation~\cite{bernardini1999ball,scheidegger2005triangulating} and deforming an initial template mesh~\cite{li2010analysis,sharf2006competing}.
Yet, these methods are sensitive to hyperparameters and initialization, requiring careful and time-consuming tuning on \emph{each} point cloud.
Some other methods~\cite{kazhdan2006poisson,kazhdan2005reconstruction,kazhdan2013screened,carr2001reconstruction,shen2004interpolating,manson2008streaming} assume that the point cloud comes with a consistently-aligned normal field.
However, it is highly non-trivial to obtain accurate orientation information for point clouds~\cite{peng2021shape,hou2022iterative}.
The final surface can be reconstructed by solving a Poisson equation~\cite{kazhdan2006poisson,kazhdan2013screened} or calculating a signed distance function using RBF~\cite{carr2001reconstruction}, moving least squares~\cite{shen2004interpolating}, Fourier coefficients~\cite{kazhdan2005reconstruction}, or wavelets~\cite{manson2008streaming}.

\vspace{-3mm}
\paragraph{Learning-based neural implicit methods.}
Recently, neural implicit function~\cite{chen2019learning,mescheder2019occupancy,park2019deepsdf} has shown its superiority in representing 3D shapes.
With the access to large shape data, such as ShapeNet~\cite{chang2015shapenet} and Thingi10k~\cite{zhou2016thingi10k},
\cite{erler2020points2surf,liao2018deep,mescheder2019occupancy,mi2020ssrnet,park2019deepsdf,badki2020meshlet,genova2020local,jiang2020local,liu2021deep} propose to learn a data prior encoded in a neural implicit function for surface reconstruction from a point cloud. 
The reconstructed surface can be obtained by applying the trained model to a new point cloud.
To enhance the generalization of the learned prior, \cite{badki2020meshlet,genova2020local,jiang2020local,liu2021deep} propose to learn local priors.
Yet, the priors learned from limited data may lead to overly-smoothed surfaces and the performance may further degrade,  if
the test point cloud is not similar to the objects in the training data.

\vspace{-3mm}
\paragraph{Fitting-based neural implicit methods.}
Instead of utilizing a data-driven prior, another stream of works attempts to leverage the neural network as a universal function approximator to solve an optimization for each point cloud input.
These methods optimize \emph{one network per object} to implicitly encode a signed distance field (SDF) whose zero-level set represents the reconstructed surface.
Some recent works propose to utilize various unsigned supervision to optimize the SDF~\cite{ atzmon2020sal,atzmon2020sald,gropp2020reg,ben2022digs}, such as unsigned distance, unsigned normal, \etc 
~Other methods~\cite{atzmon2019controlling,ma2021pull} project samples in the free space to a zero-level set and compute the distance metric with the input point cloud as the loss.
Note that we do not classify Neural-Pull~\cite{ma2021pull} as an unsigned method, since the surface cannot be represented as a zero-level set, which does not satisfy the basic property of SDF.
%
%
Our work falls into this neural optimization category and extends previous methods with a novel semi-signed supervision, a novel loss-based region sampling strategy and progressive PE, so that more complex shapes could be better reconstructed.

\begin{figure*}[tbh]
\centering
\begin{overpic}[width=1.0\textwidth]{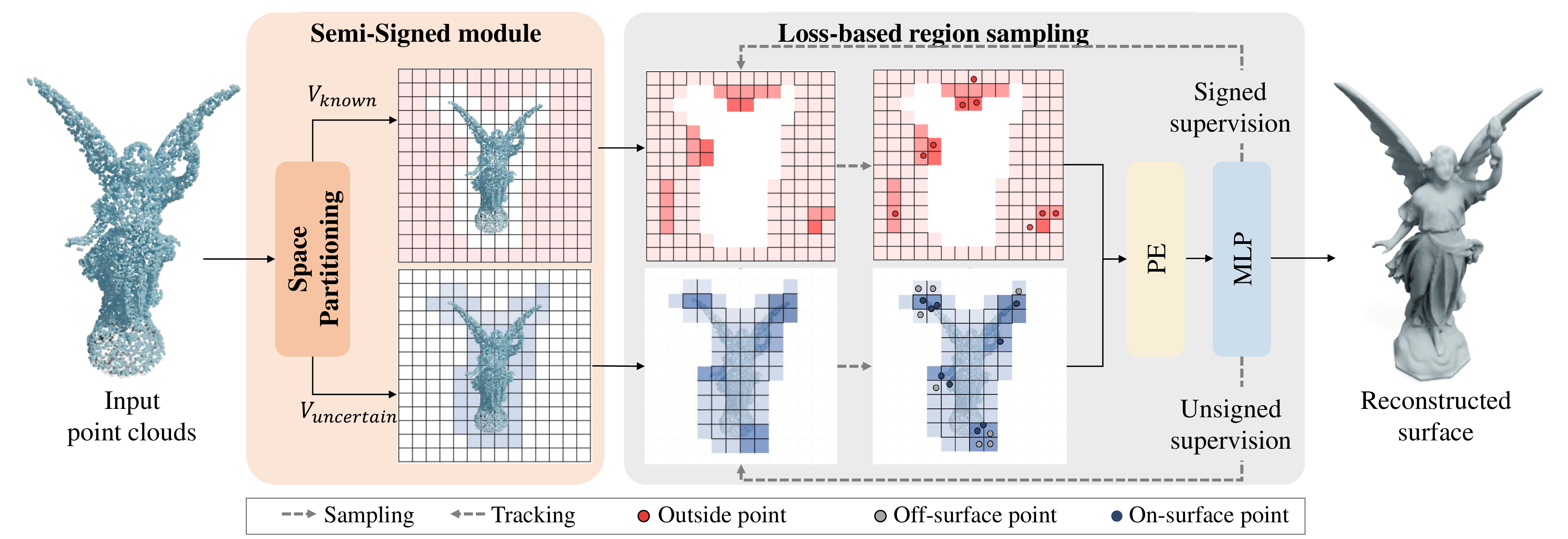}
\end{overpic}
\centering
\caption[fig:overview]{
\textbf{Overview of our SSP method.}
First, we partition the object space into ``outside'' (sign-known) regions $\mathcal{V}_{\mathrm{known}}$ and ``uncertain'' (sign-uncertain) regions $\mathcal{V}_{\mathrm{uncertain}}$, such that we can safely impose signed supervision on the outside regions.
This simple signed supervision can
effectively avoid reconstructing undesired ghost surfaces, complementing the existing unsigned supervisions (Sec.~\ref{sec:semi_signed}).
Second, to reconstruct fine structures better, we propose a loss-based region sampling strategy (Sec.~\ref{sec:sampling}) to adaptively increase the sampling frequency in complicated regions with larger losses and utilize progressive positional encoding (Sec.~\ref{sec:pe}) for fitting high-frequency signal.
}
\label{fig:overview}
\end{figure*}

\section{Method} 
\label{sec:method}
Given an \emph{unoriented} point cloud $\mathcal{P}$, our objective is to optimize a network $f_\theta$ (parameterized by $\theta$) to reconstruct the underlying surface represented with a signed distance field (SDF).
Then, we can obtain the explicit surface $S_\theta$
by extracting the zero-level set from $f_\theta$:
\begin{equation}
\setlength{\abovedisplayskip}{3pt}  
\setlength{\belowdisplayskip}{3pt}  
\label{SDF}
S_\theta = \{ p\in \mathbb{R}^3 | f_\theta(p)=0 \}.
\end{equation}
To obtain an accurate surface, \emph{appropriate supervisions} are required to guide the optimization for the network $f_\theta$.
Specifically, the designed supervisions are applied either on the \emph{on-surface} samples (denoted as $p \in \mathcal{P}$) or on the \emph{off-surface} samples (denoted as $q \in \mathbb{R}^3 - \mathcal{P}$)
during the optimization process.
The on-surface losses are used to encourage faithful reconstruction on the sampled surface points, while the off-surface losses are used 
as regularization to suppress the existence of degenerated
structures.
%
%

%
\subsection{Revisiting unsigned supervision}
\label{sec:revisit}
Prior to our improvements in Sec.~\ref{sec:semi_signed}-\ref{sec:pe},
we first revisit existing supervisions (losses) and discuss their limitations. 
The existing supervisions can be categorized into on-surface and off-surface supervisions.

\vspace{-5mm}
\paragraph{On-surface supervision.}
On-surface distance loss $\mathcal{L}^\mathrm{on}_{\mathrm{dist}}$~\cite{gropp2020reg} 
encourages the extracted zero-level set contains the existing points in given point clouds: 
\begin{equation}
\setlength{\abovedisplayskip}{3pt}  
\setlength{\belowdisplayskip}{3pt}  
\label{eq:dist-on}
\mathcal{L}^\mathrm{on}_{\mathrm{dist}}=\sum\limits_{p \in \mathcal{P}} ||f_\theta(p)||.
%
\end{equation} 
The on-surface unoriented derivative loss $\mathcal{L}^\mathrm{on}_{\mathrm{grad}}$~\cite{atzmon2020sald} 
constrains $\nabla f_\theta(p)$ with the given unoriented surface normal $n_p$ at $p$:
%
%
\begin{equation}
\setlength{\abovedisplayskip}{3pt}  
\setlength{\belowdisplayskip}{3pt}  
\label{eq:SALD}
\mathcal{L}^\mathrm{on}_{\mathrm{grad}} = \sum\limits_{p \in \mathcal{P}} min\{||\nabla f_\theta(p)-n_p||,||\nabla f_\theta(p)+n_p||\},
\end{equation}
where $\nabla f_\theta(p) $ is the derivative of the network in 
point $p$.

\vspace{-3mm}
\paragraph{Off-surface supervision.}
SAL~\cite{atzmon2020sal} supervises the SDF predictions $f_\theta(q)$ with unsigned distance $d$, approximated with unsigned distance from $q$ to its closest point in $\mathcal{P}$:
\begin{equation}
\setlength{\abovedisplayskip}{3pt}  
\setlength{\belowdisplayskip}{3pt}  
\label{eq:SAL}
\mathcal{L}^\mathrm{free}_{\mathrm{dist}} = \sum\limits_{q \in \mathbb{R}^3 \setminus \mathcal{P}}min\{||f_\theta(q)-d||,||f_\theta(q)+d||\}.
\end{equation}
Besides, IGR~\cite{gropp2020reg} utilizes the Eikonal regularization~\cite{crandall1983viscosity}, which encourages SDF to maintain unit-length gradients in the whole space to produce a valid SDF. 
\begin{equation}
\setlength{\abovedisplayskip}{3pt}  
\setlength{\belowdisplayskip}{3pt}  
\label{eq:IGR}
\mathcal{L}_\mathrm{E} = \sum\limits_{q \in \mathbb{R}^3 \setminus \mathcal{P}}(\left \|  \nabla f_\theta(q) - 1 \right \|)^2.
\end{equation}

\vspace{-5mm}
\paragraph{Discussion.}

Although methods using the existing unsigned supervision demonstrate promising reconstructions from raw point clouds, we notice that their success is highly dependent on whether the sphere initialization~\cite{gropp2020reg} is a good ``coarse shape'' of the target point clouds.
For example, the initial sphere is not a good approximation of the target point clouds when there exists a huge volume difference between them (even if the point cloud is enclosed in the sphere).
Indeed, most existing methods (\eg, IGR~\cite{gropp2020reg}, SAL~\cite{atzmon2020sal} and SALD~\cite{atzmon2020sald}) tend to fail for these shapes; see, \eg, Fig.~\ref{fig:teaser}(a), Fig.~\ref{fig:result_abc}, and Fig.~\ref{fig:result_density}. More examples are provided in the Supp.)
This observation motivates us to determine a rough \emph{outside} region and apply more informative signed supervision on it, so that the network can better avoid bad local minimums.

%
%
%


%

\subsection{Semi-signed optimization module}
\label{sec:semi_signed}
To alleviate the difficulty in finding a coarse shape, we propose a simple yet effective solution that utilizes more informative signed supervision in regions apparently \emph{outside} of the object.
More concretely, the underlying surface of a point cloud is bounded, and the sign of regions outside this bounding hull should be positive.
We propose the \emph{semi-signed optimization module}, that automatically finds outside regions and applies more informative signed supervision.

\paragraph{Space partitioning.}
To apply signed supervision, we first need to 
find the apparently ``outside'' region $\mathcal{V}_{\mathrm{known}}$.
The complement region of $\mathcal{V}_{\mathrm{known}}$ is the \emph{sign-uncertain} region, in which the signs of its points are not known beforehand.
Note that this partition does not need to be precise to help the network quickly find a coarse shape during the optimization. 

More concretely, the input point cloud is first normalized to a cube ranged $[-0.9, +0.9]^3$, following the common practice in~\cite{gropp2020reg,atzmon2020sal,atzmon2020sald}.
Then, we voxelize the space $[-1.0, +1.0]^3$ into an $N^3$ grid, where $N$ is the resolution calculated based on the density of input point clouds (see Supp. for the details). 
For the boundary voxels, it is easy to determine whether they belong to the \emph{outside} region by simply testing if it contains any point in the given cloud points.
Starting from the \emph{outside} boundary voxels, we recursively use a breadth-first search (BFS) 
to find outside voxels connected to them.
The search stops if a voxel or any of its neighboring voxels contains a point in the given point cloud (\ie, approaching the neighboring of object boundary).
The set of empty voxels forms $\mathcal{V}_{\mathrm{known}}$ after the recursive search.
The pseudo code for this procedure is provided in Supp.
%

\begin{figure}[tb]
\begin{center}
\begin{overpic}[width=0.5\textwidth]{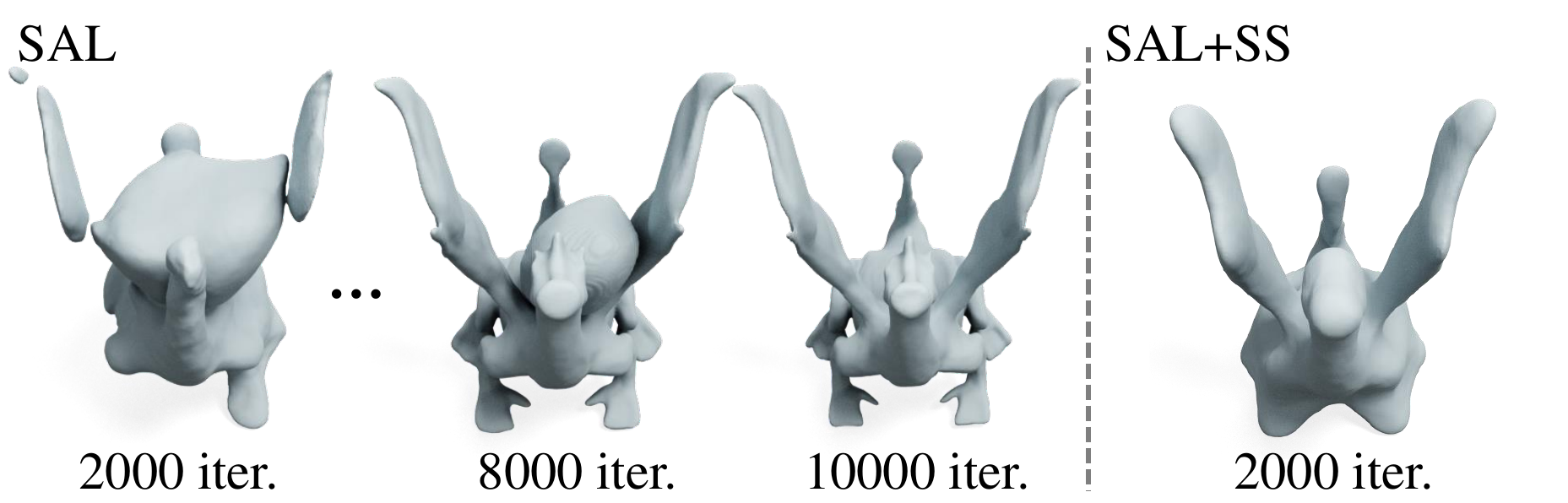}%
\end{overpic}
\end{center}
\vspace{-2mm}
\caption{
Visual comparison of the intermediate results optimized by SAL~\cite{atzmon2020sal} and ``SAL + SS'', where SS means signed supervision.
}
\label{fig:intermediate_result}
\end{figure}

\vspace{-5mm}
\paragraph{Signed supervision in sign-known regions.}
We enforce the signs predicted by $f_\theta(q), q \in \mathcal{V}_{\mathrm{known}}$ to be positive.
Mathematically, we propose the following loss function: 
\begin{equation}
\setlength{\abovedisplayskip}{3pt}  
\setlength{\belowdisplayskip}{3pt}  
\label{eq:signed_loss}
\mathcal{L}_{\mathrm{signed}} = \sum\limits_{q \in \mathcal{V}_{\mathrm{known}}}\tau(\epsilon-f_{\theta}(q))_{+},
\end{equation}
where $\epsilon=\frac{1}{N}$ is a positive margin distance and $\tau(x)_{+}:=max(x,0)$. 
According to Eq.~\eqref{eq:signed_loss}, this loss only imposes the penalization when the predicted SDF value in $\mathcal{V}_{\mathrm{known}} $ is smaller than $\epsilon$.
Although we do not use the exact signed distance as supervision, we observe that this signed supervision can help the network quickly learn a coarse shape of the target point cloud, compared with existing unsigned methods; see an example in Fig.~\ref{fig:intermediate_result}.
\vspace{-3mm}
\paragraph{Unsigned supervision in sign-uncertain region.}
For the $\mathcal{V}_{\mathrm{uncertain}}$ region, 
only unsigned supervisions~\cite{gropp2020reg,atzmon2020sal,atzmon2020sald} can be used.
Specifically, we adopt both 
distance loss $\mathcal{L}^\mathrm{on}_{\mathrm{dist}}$ (Eq.~\eqref{eq:dist-on}) 
and derivative loss $\mathcal{L}^\mathrm{on}_{\mathrm{grad}}$ (Eq.~\eqref{eq:SALD}) 
for on-surface points, 
$\mathcal{L}^\mathrm{free}_{\mathrm{dist}}$ (Eq.~\eqref{eq:SAL}) 
, and Eikonal regularization $\mathcal{L}_\mathrm{E}$ (Eq.~\eqref{eq:IGR}) 
for off-surface points.
To add more constraints on the $\mathcal{V}_{\mathrm{uncertain}}$ region, we introduce an extra first-order guidance (\ie, unsigned derivative loss) for $q$ as follows:
\begin{equation}
\setlength{\abovedisplayskip}{3pt}  
\setlength{\belowdisplayskip}{3pt}  
\label{eq:extend-derivative}
\mathcal{L}^\mathrm{free}_{\mathrm{grad}} = \sum\limits_{q \in \mathcal{V}_{\mathrm{uncertain}}\setminus \mathcal{P}} min\{||\nabla f_\theta(q)-n_p||,||\nabla f_\theta(q)+n_p||\},
\end{equation}
where $p\in\mathcal{P}$ is the nearest point to $q$.
Here, we utilize the unoriented normal of $p$ to approximate the derivative on its neighboring point $q$ as a regularization.
Please refer to Supp. for our motivation and more discussion.
%
%
%

\vspace{-5mm}
\paragraph{Overall optimization objective.}
The overall loss is the weighted sum of the aforementioned loss terms:
\begin{equation}
\setlength{\abovedisplayskip}{3pt}  
\setlength{\belowdisplayskip}{3pt}  
\label{eq:full-loss}
\begin{split}
\mathcal{L} = & w_1 \mathcal{L}^\mathrm{on}_{\mathrm{dist}} + w_2 \mathcal{L}^\mathrm{free}_{\mathrm{dist}} + w_3 \mathcal{L}^\mathrm{on}_{\mathrm{grad}} + \\
& w_4 \mathcal{L}^\mathrm{free}_{\mathrm{grad}} + w_5 \mathcal{L}_\mathrm{E} + w_6 \mathcal{L}_{\mathrm{signed}},
\end{split}
\end{equation}
where $\{w_i\}$ are weights of each loss term.

\subsection{Loss-based per-region sampling} \label{sec:sampling}
We propose to sample more points for optimization, if a region has a large tracked loss.
This strategy can be seen as a new variant of importance sampling
and can bring two main benefits: 
(i) facilitates the reconstruction of fine details in complicated regions, as regions with more details are normally harder to fit; and
(ii) helps avoid creating degenerated surfaces in free space when applied together with the signed supervision. Note that our sampling strategy is also used in the outside region.
Procedure-wise, it has two main steps: region-wise loss tracking and adaptive sampling.

\vspace{-5mm}
\paragraph{Region-wise loss tracking.}
We consider all five major loss terms: $\mathcal{L}^\mathrm{on}_{\mathrm{dist}}$, $\mathcal{L}^\mathrm{free}_{\mathrm{dist}}$, $\mathcal{L}^\mathrm{on}_{\mathrm{grad}}$, $\mathcal{L}^\mathrm{free}_{\mathrm{grad}}$, $\mathcal{L}_{\mathrm{signed}}$.
Taking $\mathcal{L}^{on}_{\mathrm{dist}}$ as an example, we track the running mean loss of all applicable voxels $\mathcal{V}^{i}$ (occupied voxels for $\mathcal{L}^{on}_{\mathrm{dist}}$) as follows:
\begin{equation}
\setlength{\abovedisplayskip}{5pt}  
\setlength{\belowdisplayskip}{5pt}  
\label{eq:update_moving}
\mathcal{M}^{i-\mathrm{on}}_{\mathrm{dist}} = (1 - \alpha) \times \mathcal{M}^{i-\mathrm{on}}_{\mathrm{dist}} + \alpha \times \mathcal{L}^{i-\mathrm{on}}_{\mathrm{dist}},
\end{equation}
\noindent where 
$\mathcal{L}^{i-\mathrm{on}}_{\mathrm{dist}}$ is the loss of voxel $i$ in the current iteration,
$\mathcal{M}^{i\mathrm{-on}}_{\mathrm{dist}}$ is the tracked running mean loss, 
and $\alpha$ is the momentum empirically set to 0.1.
Similarly, we can track the other losses denoted as $\mathcal{M}^{i\mathrm{-on}}_{\mathrm{grad}}$,
$\mathcal{M}^{i\mathrm{-free}}_{\mathrm{dist}}$,
$\mathcal{M}^{i\mathrm{-free}}_{\mathrm{grad}}$, 
and $\mathcal{M}^{i}_{\mathrm{signed}}$.
Note that different losses are applied to different regions (voxels).
Please refer to Supp. for more details.

\vspace{-5mm}
\paragraph{Adaptive sampling.}
Next, we perform a two-step sampling for each loss: (i) adaptively sample a set of voxels based on the previous region losses, and (ii) sample points within each region (voxels). 
In the first step, we sample a set of voxels, where each voxel $\mathcal{V}^{i}$ is assigned with a sampling probability $p^{i}_{\mathrm{type}}$ proportional to its tracked moving average loss:
\begin{equation}
\setlength{\abovedisplayskip}{3pt}  
\setlength{\belowdisplayskip}{3pt}  
\label{sampling}
p_{\mathrm{type}-i} = \frac{ \mathcal{M}_{\mathrm{type}-i}}{\sum_j{\mathcal{M}_{\mathrm{type}-j}}},
\end{equation} 
where $\mathcal{M}_{\mathrm{type}-j} \in \{\mathcal{M}^{j\mathrm{-on}}_{\mathrm{dist}}, \mathcal{M}^{j\mathrm{-on}}_{\mathrm{grad}}, \mathcal{M}^{j\mathrm{-off}}_{\mathrm{dist}}, \mathcal{M}^{j\mathrm{-off}}_{\mathrm{grad}}, \allowbreak \mathcal{M}^{j}_{\mathrm{signed}}\} $.
Then, we apply different procedures to obtain the final point samples according to the loss type as follows.

\begin{figure}[t]
\begin{center}
\begin{overpic}[width=0.45\textwidth]{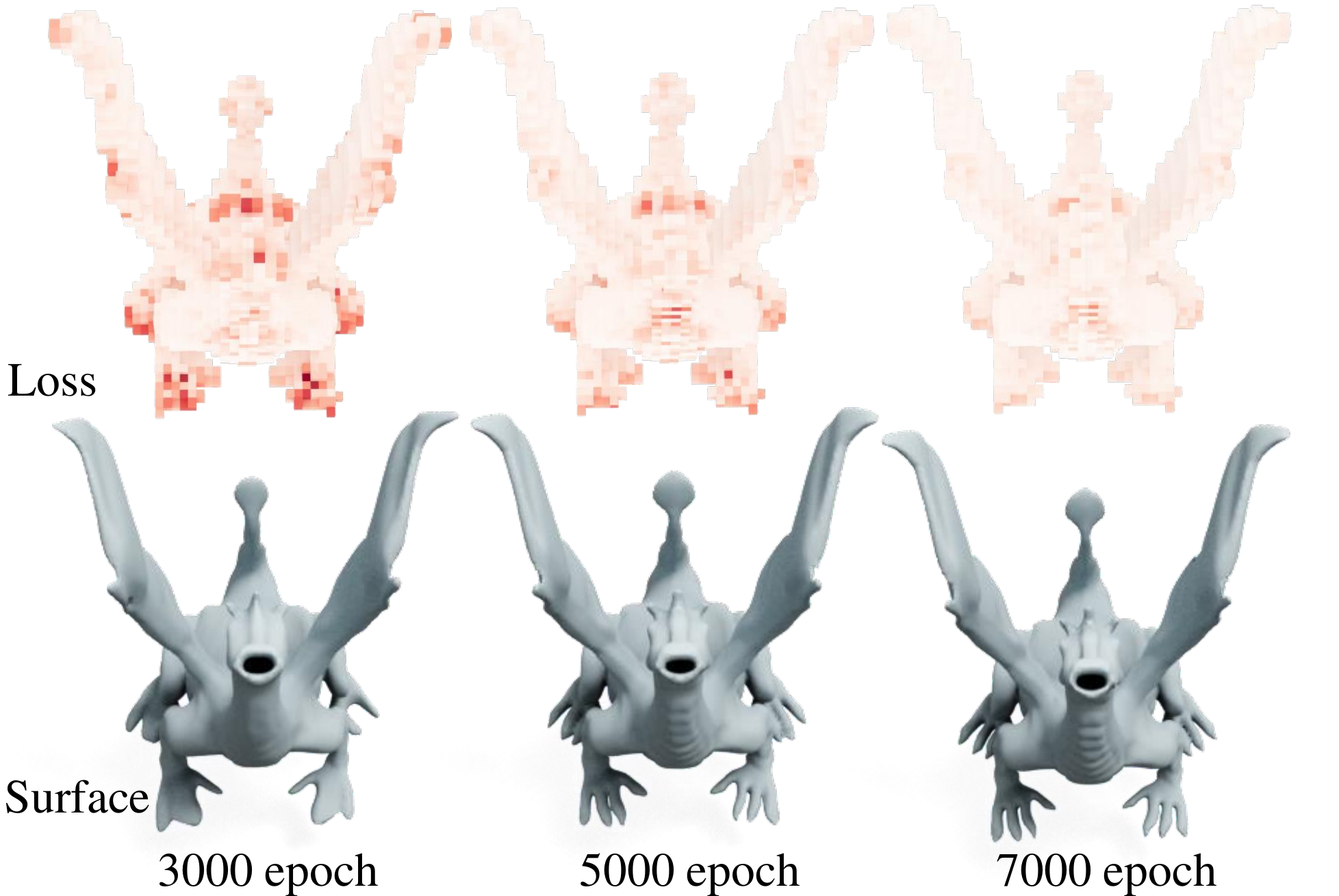}
\end{overpic}
\end{center}
\vspace{-3mm}
\caption{
Visualizing the tracked moving average loss ($\mathcal{M}^{j\mathrm{-on}}_{\mathrm{grad}}$) in every voxel (top) and reconstructed surface (bottom) at different iterations.
Deeper red colors indicate voxels of higher losses.
}
\label{fig:loss-progress}
\end{figure}

For off-surface losses ($\mathcal{L}^\mathrm{off}_{\mathrm{dist}}$, $\mathcal{L}^\mathrm{off}_{\mathrm{grad}}$) and the signed loss $\mathcal{L}_{\mathrm{signed}}$, 
we use a uniform random sampling strategy to sample the points within the voxel.
For the on-surface losses ($\mathcal{L}^\mathrm{on}_{\mathrm{dist}}$ and $\mathcal{L}^\mathrm{on}_{\mathrm{grad}}$), we query the k-nearest neighbor (k-NN) input points to the voxel center.
Note that a small Gaussian noise is added to the center location before k-NN, so that different points can be sampled in different iterations.
Fig.~\ref{fig:loss-progress} shows a visualization of the sampling probabilities of the voxels. 
Note that our strategy tends to sample regions with complicated structures and fine details. 
The ablation study in Sec.~\ref{ablation} (Fig.~\ref{fig:visualtization-ablation}) also shows that our proposed sampling strategy helps the network better fit the complicated regions.
%
%
%
More details are provided in Supp.

\subsection{Progressive Positional Encoding}
\label{sec:pe}

Inspired by the recent works~\cite{mildenhall2021nerf,hertz2021sape, wang2022improved, park2021nerfies}, we also adopt a kind of progressive positional encoding (PE) to gradually introduce high-frequency components into the PE to enhance the optimization quality.
The progressive strategy is particularly helpful, especially when there exist complicated structures and fine details.
Specifically, the progressive positional encoding can be represented by
\begin{equation}
\setlength{\abovedisplayskip}{5pt}  
\setlength{\belowdisplayskip}{5pt}  
\begin{aligned}
\gamma(\mathbf{p})=
&  \left(M_{0}\sin \left(2^0 \pi \mathbf{p}\right), 
M_{0}\cos \left(2^0 \pi \mathbf{p}\right), 
\ldots\right. \\
& M_{L-1}\left.\sin \left(2^{L-1} \pi \mathbf{p}\right), 
M_{L-1}\cos \left(2^{L-1} \pi \mathbf{p}\right)\right),
\end{aligned}
\end{equation}
where $\mathbf{p}$ is a point coordinate,
$L$ is the frequency band, 
and $\mathbf{M}=\{M_{0},\cdots, M_{L-1}\}$ is a mask vector.
We use the following progressive rule for $M_{i}$ at iteration $n$:
\begin{equation}  
\setlength{\abovedisplayskip}{5pt}  
\setlength{\belowdisplayskip}{5pt}  
M_{i}=\left \{
\begin{array}{ll}
0, & \text { if } i > L_{0} + \frac{n}{k} \\
1, & \text { otherwise },
\end{array} \right.
\end{equation}
where $i \in \{0, 1, ..., L-1\}$, 
$k$ is a parameter controlling the increasing speed,
and $L_{0}$ is the initial frequency band.
We empirically set $L_{0}=3, L=6$, and $k=1000$. 


\section{Results}
\label{sec:result}

We conducted extensive experiments on
the ABC subset (100 objects)~\cite{erler2020points2surf}, 
and three types of challenging data that contain objects with fine details~\cite{zhou2016thingi10k} (5 objects), 
varying sampling density (32 objects)~\cite{guerrero2018pcpnet}, 
and noises (57 objects)~\cite{guerrero2018pcpnet}. 

\subsection{Experiment setting}
\paragraph{Evaluation metrics.}
We report the L1-based Chamfer Distance (denoted as CD-$L_1$), Normal Consistency (NC), and F1-score with default threshold $1\%$ following SAP~\cite{peng2021shape} (F-score) between reconstructed surface and GT.
Following SAP~\cite{peng2021shape}, we use the same procedure as  SAP~\cite{peng2021shape} to calculate the metrics (details provided in Supp.).
\vspace{-5mm}
\paragraph{Baselines.}
We compare with three types of methods:
(i) non-neural implicit methods including Screened Poisson Surface Reconstruction (SPSR)~\cite{kazhdan2013screened} and SAP~\cite{peng2021shape};
(ii) optimization-based neural implicit methods, including IGR~\cite{gropp2020reg}, SAL~\cite{atzmon2020sal}, SALD~\cite{atzmon2020sald}, Neural-Pull~\cite{ma2021pull}, and DiSG~\cite{ben2022digs}; 
and (iii) a learning-based neural implicit method, IMLSNet~\cite{liu2021deep}.
The results of IMLSNet~\cite{liu2021deep} (denoted as IMLS
) are calculated based on their officially-released pre-trained model.
For the other baselines (IGR~\cite{gropp2020reg}, SAL~\cite{atzmon2020sal}, Neural-Pull~\cite{ma2021pull} (denoted as N-P) and SAP~\cite{peng2021shape}), 
we report the results optimized by their officially-released code.
The normal inputs of our method, SALD~\cite{atzmon2020sald}, and SPSR~\cite{kazhdan2013screened} are estimated from the state-of-the-art method AdaFit~\cite{zhu2021adafit}.
Since SPSR~\cite{kazhdan2013screened} requires consistently oriented normals, we use minimum spanning tree~\cite{zhou2018open3d} to propagate the normal orientations~\cite{hoppe1992surface}.

\vspace{-4mm}
\paragraph{Implementation details.}
We use the same neural architecture and the same number of sampling points as IGR~\cite{gropp2020reg}.
Following~\cite{atzmon2020sal,atzmon2020sald,gropp2020reg,ma2021pull}, we overfit the network for every input point cloud and set the full iteration as 10,000 for the whole experiments.
We adopt the Marching cubes~\cite{lorensen1987marching} algorithm to extract the mesh from its implicit field, and the default resolution is $256^3$ 
for all the neural implicit methods.
All the experiments are conducted on the GeForce RTX 2080 Ti GPU.
Details and architectures can be found in Supp. 
\begin{figure}[tb]
\begin{center}
    \begin{overpic}[width=1.0\linewidth]{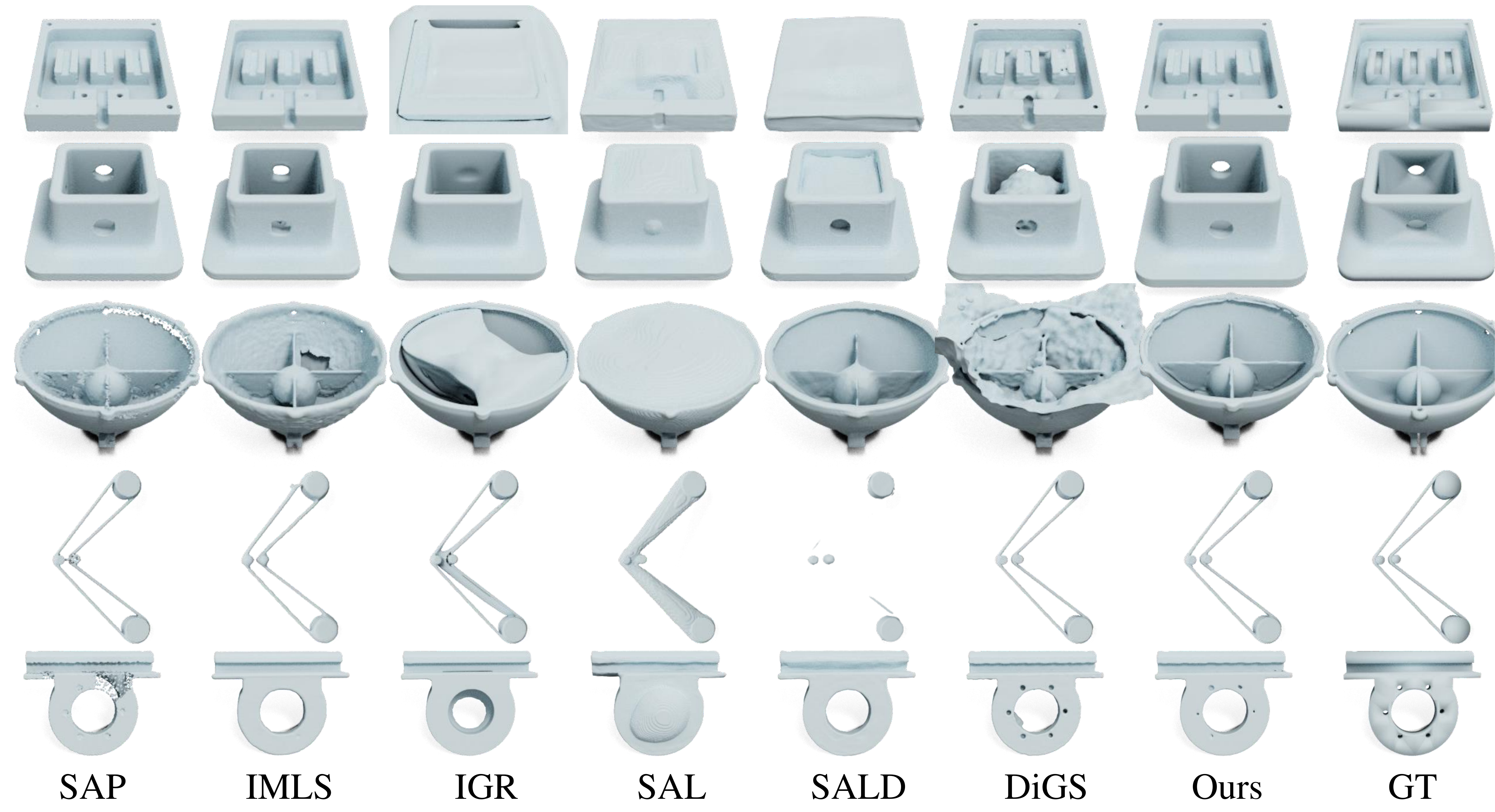}%
\end{overpic}
\end{center}
\vspace{-1pt}
\caption{
Visual comparisons on the ABC subset~\cite{erler2020points2surf}.
}
\label{fig:result_abc}
\end{figure}
\subsection{Result on ABC dataset}
\begin{table}[tb]

\begin{center}
\small
{\def\arraystretch{1} \tabcolsep=1.5em 
\begin{tabular}{lccc}
\toprule
Methods &
  F-score$_\uparrow$ &
  CD-$L_1$ $_\downarrow$ &
  NC$_\uparrow$ \\ 
\hline
SPSR~\cite{kazhdan2013screened} & 0.557     & 2.774 & 0.904  \\
SAP~\cite{peng2021shape} & 0.660      & 1.368 & 0.915  \\
\hline
IMLS~\cite{liu2021deep} & 0.626     & 1.245 & 0.923   \\
\hline
N-P~\cite{ma2021pull} & 0.370      & 2.071 & 0.912 \\
SAL~\cite{atzmon2020sal} & 0.407 & 4.676 & 0.870 \\
SALD~\cite{atzmon2020sald} & 0.560 & 1.719 & 0.919 \\
IGR~\cite{gropp2020reg} & 0.551     & 4.429 & 0.891  \\
DiGS~\cite{ben2022digs}  & 0.657 & 1.540 & 0.936\\
\hline
Ours  & \textbf{0.675}     & \textbf{1.225} & \textbf{0.938}   \\
\bottomrule
\end{tabular}
}
\end{center}
\vspace{-3mm}
\caption{
Quantitative comparisons on the ABC subset~\cite{erler2020points2surf}. 
}
\label{Tab:abc_metric}
\end{table}
To evaluate the stability of our method, we compare our method in a subset of ABC~\cite{koch2019abc} (noise-free version) released by Points2surf~\cite{erler2020points2surf}.
This subset contains 100 
3D CAD models with various topologies
and we use the provided sampled point clouds as input.
Quantitative results of different methods are shown in Tab.~\ref{Tab:abc_metric}.
Overall, our method achieves the best results on all the metrics (\ie, F-score, CD-$L_1$, NC).
The significant decrease in CD-$L_1$ demonstrates that our method could effectively avoid generating ghost surfaces on various shapes.
As shown in Fig.~\ref{fig:result_abc}, our method could significantly alleviate the ghost surface problem and reconstruct more accurate surfaces.
In contrast, most other optimization-based methods 
(IGR~\cite{gropp2020reg}, SAL~\cite{atzmon2020sal}, SALD~\cite{atzmon2020sald} and DiGS~\cite{ben2022digs}) 
occasionally create ghost surfaces in undesired locations, thereby severely worsening the CD-$L_1$ value.
Note that the surface reconstructed by SAP~\cite{peng2021shape} contains small artifacts on the edge regions (3rd row in Fig.~\ref{fig:result_abc}).
IMSL~\cite{liu2021deep} reconstructs accurate surfaces for simple objects but suffers from inaccurate predictions on thin structures (3rd row in Fig.~\ref{fig:result_abc}).
More visual comparisons can be found in Supp.

\subsection{Result on challenging data}

\paragraph{Data with high-level details}
\begin{table}[tb]
\begin{center}
\small
\setlength{\tabcolsep}{1.0em}{
\begin{tabular}{lccc}
\toprule
Methods &   F-score$_\uparrow$ &  CD-$L_1$ $_\downarrow$ &   NC$_\uparrow$   \\ 
\hline
SPSR~\cite{kazhdan2013screened} & 0.921 & 0.744 & 0.955  \\
SAP~\cite{peng2021shape}     & 0.936     & 0.536    & 0.937   \\\hline
IMLS~\cite{liu2021deep} & 0.793     & 0.759    & 0.882  \\\hline
N-P~\cite{ma2021pull}      & 0.627     & 0.934    & 0.927     \\
SAL~\cite{atzmon2020sal} & 0.884 & 0.779 & 0.925 \\
SALD~\cite{atzmon2020sald} & 0.730 & 1.187 & 0.891  \\
IGR~\cite{gropp2020reg} & 0.308 & 6.471 & 0.631  \\
DiGS~\cite{ben2022digs}    & 0.942 & 0.529 & 0.954    \\
\hline
Ours  & \textbf{0.943} & \textbf{0.520} & \textbf{0.960} \\
\bottomrule
\end{tabular}
}
\end{center}
\vspace{-3mm}
\caption{
Quantitative comparisons on complex shapes in Thingi10K~\cite{zhou2016thingi10k}.
}
\label{Tab:challenging}
\end{table}

We further conduct experiments on Thingi10k~\cite{zhou2016thingi10k}.
Similar to prior work~\cite{peng2021shape}, we use the same five challenging shapes with complex topology and high-level fine details.
As shown in Tab.~\ref{Tab:challenging}, we achieve the best performance on all metrics on Thingi10k~\cite{zhou2016thingi10k}.
As shown in Fig.~\ref{fig:thingi10k_srb_data}, our method could preserve more details compared with existing methods while having fewer artifacts.
Although DiGS~\cite{ben2022digs} uses the SIRENs~\cite{sitzmann2020implicit} neural structure for high-frequency representation, it still has difficulty in fitting the fine structures, since the points in these regions have not been adaptively sampled as many times as ours due to its uniform sampling strategy.
See the abdomen of the ``dinosaur'' shown in Fig.~\ref{fig:thingi10k_srb_data}.
Though SAP~\cite{peng2021shape} could successfully reconstruct an accurate surface, it occasionally creates artifacts for the complicated region.
\begin{figure}[tb]
\begin{center}
\begin{overpic}[width=1.0\linewidth]{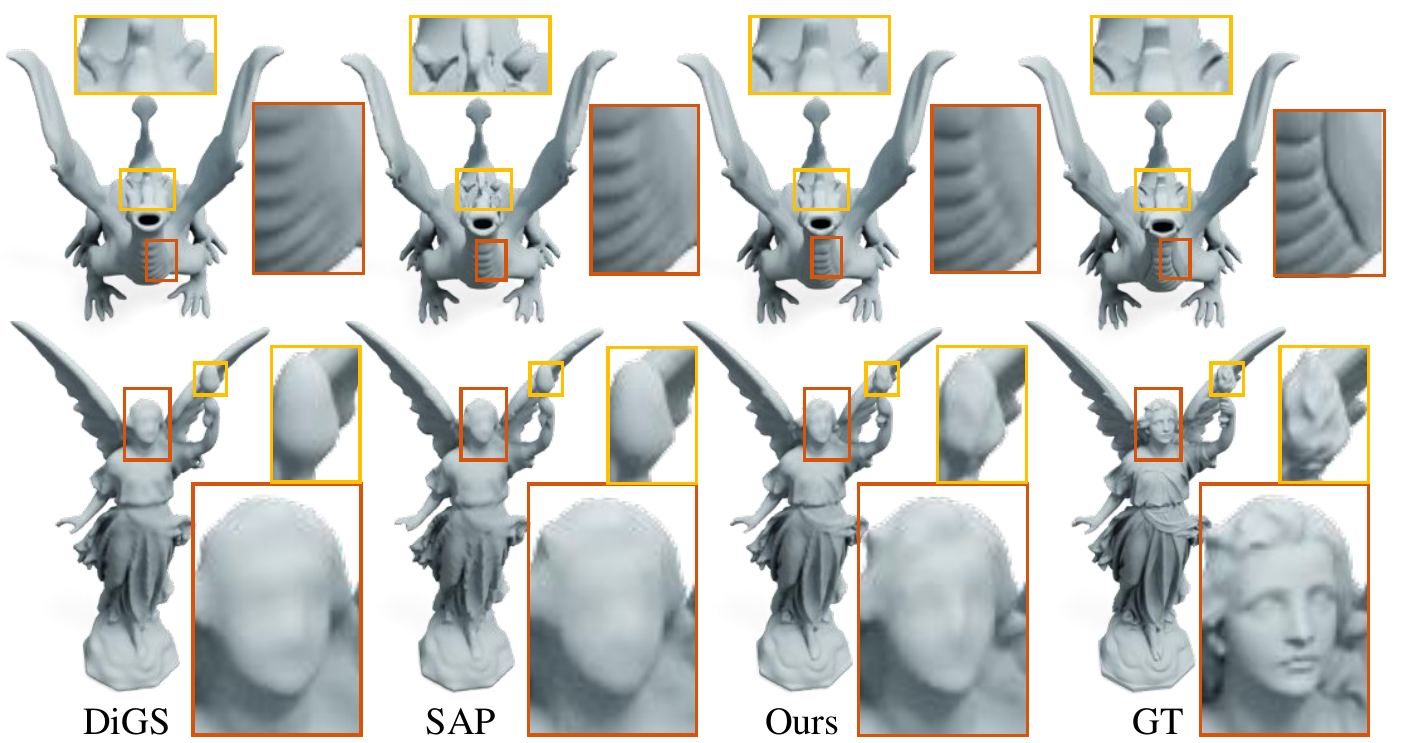}
\end{overpic}
\end{center}
\vspace{-3mm}
\caption{Visual comparisons on Thingi10K~\cite{zhou2016thingi10k}.
}
\label{fig:thingi10k_srb_data}
\end{figure}

\paragraph{Varying-density data}
\begin{figure}[tb]
\begin{center}
\begin{overpic}[width=1.0\linewidth]{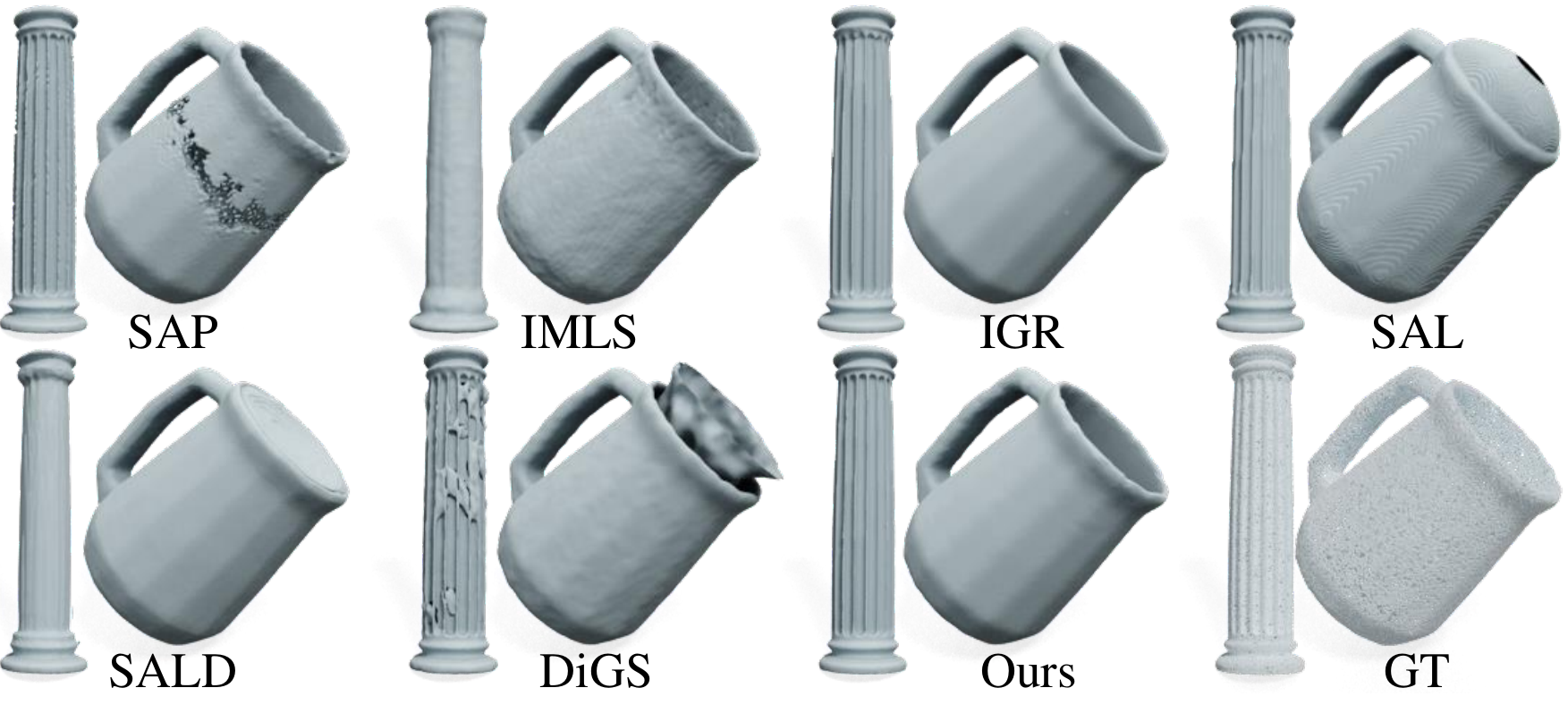}
\end{overpic}
\end{center}
\vspace{-3mm}
\caption{
Visual comparisons on varying-density data from \cite{guerrero2018pcpnet}.
}
\label{fig:result_density}
\end{figure}

\begin{table}[tb]

\begin{center}
\small
{\def\arraystretch{1} \tabcolsep=0.07em 
\begin{tabular}{lcccccc}
\toprule
&  \multicolumn{3}{c}{Density-variation} & \multicolumn{3}{c}{Noise} \\
\cmidrule(lr){2-4}\cmidrule(lr){5-7} 
Methods &
  F-score$_\uparrow$ &
  CD-$L_1$ $_\downarrow$  &
  NC$_\uparrow$ &
  F-score$_\uparrow$ &
  CD-$L_1$ $_\downarrow$ &
  NC$_\uparrow$ \\ 
\hline
SPSR~\cite{kazhdan2013screened}  & 0.789     & 2.007 & 0.938  & \textbf{0.723}     & 2.216 & 0.833 \\
SAP~\cite{peng2021shape} & 0.889     & 0.658 & 0.932 & 0.580      & 1.128 & 0.693 \\
\hline
IMLS~\cite{liu2021deep} & 0.830      & 0.715 & 0.925  & 0.583     & 1.205 & 0.879 \\
\hline
N-P~\cite{ma2021pull}  & 0.397     & 1.359 & 0.945 & 0.257     & 2.027 & 0.901 \\
SAL~\cite{atzmon2020sal}  & 0.767 & 1.823 & 0.937 & 0.328 & 8.467 & 0.88 \\
SALD~\cite{atzmon2020sald}  & 0.724 & 1.209 & 0.926 & 0.255 & 3.472 & 0.919 \\
IGR~\cite{gropp2020reg} & 0.714     & 7.316 & 0.918 & 0.697     & 3.480  & 0.889 \\
DiGS~\cite{ben2022digs}   & 0.877 & 0.868 & 0.951 & 0.544 & 1.273 & 0.717 \\
\hline
Ours  & \textbf{0.917}     & \textbf{0.567} & \textbf{0.962} & 0.685 & \textbf{0.994} & \textbf{0.957}  \\
\bottomrule
\end{tabular}
}
\end{center}
\vspace{-3mm}
\caption{
Quantitative comparisons on the data with different levels of noise and different density variation on PCPNet~\cite{guerrero2018pcpnet}.
}
\label{Tab:noisy_density_metric}
\end{table}

To evaluate the robustness of our method to varying sampling density, 
we conduct experiments on the varying-density dataset released by PCPNet~\cite{guerrero2018pcpnet}. 
As shown in Tab.~\ref{Tab:noisy_density_metric}, our method performs the best on all three metrics.
As shown Fig.~\ref{fig:result_density}, 
our method produces more complete surfaces without ghost surfaces compared with the baseline methods.
More results can be found in Supp.
\paragraph{Noisy data}
\begin{figure}[tb]
\begin{center}
\begin{overpic}[width=1.0\linewidth]{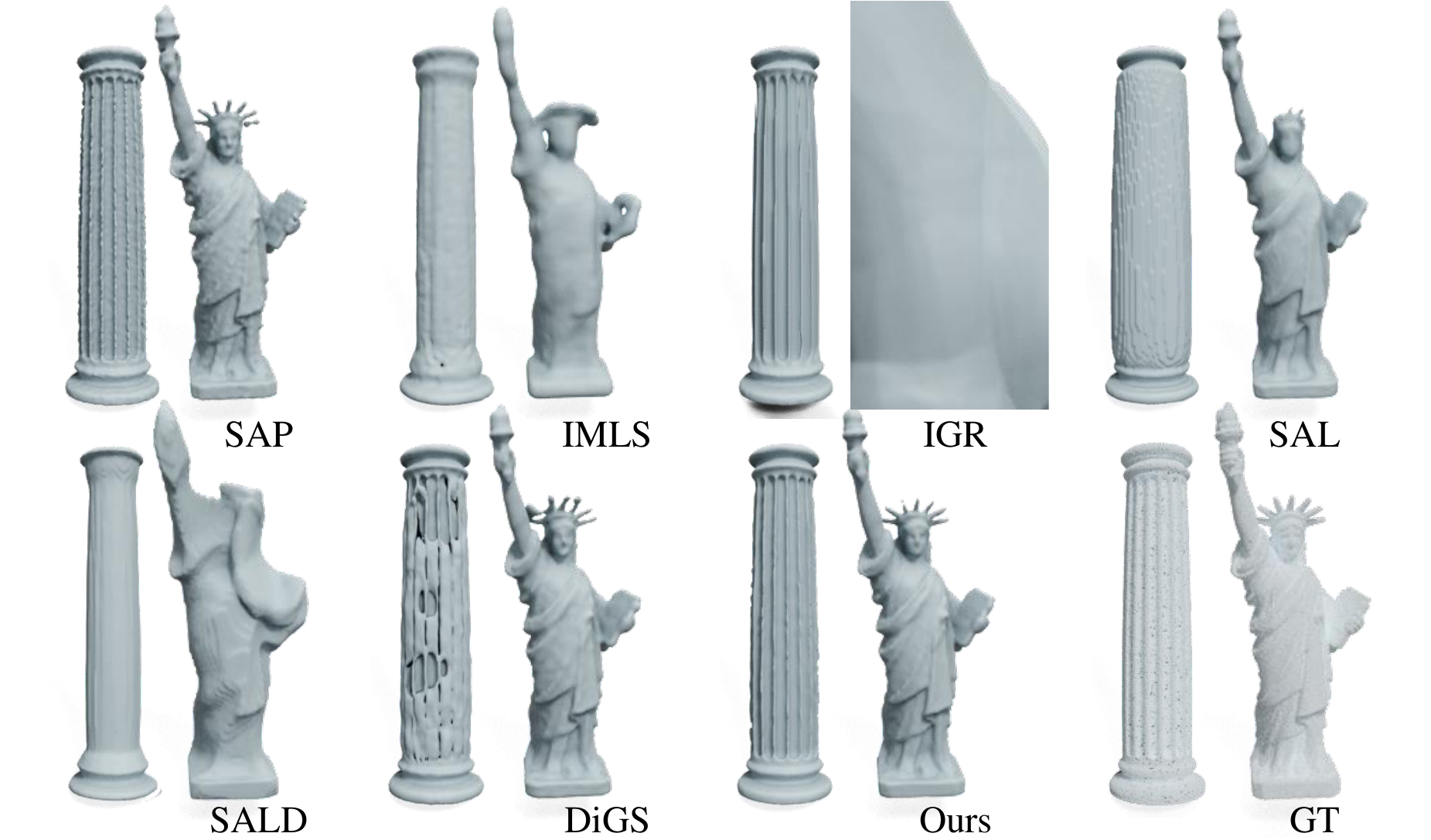}
\end{overpic}
\end{center}
\caption{
Visual comparisons on noisy data from PCPNet~\cite{guerrero2018pcpnet}.
}
\label{fig:results_noisy}
\end{figure}
To test the robustness of our method to noise, we conduct experiments on the noisy test data from PCPNet~\cite{guerrero2018pcpnet}.
This dataset includes 57 point cloud shapes of different levels of Gaussian noise.
Our method significantly outperforms others on the CD-$L_1$ and NC metrics; see Tab.~\ref{Tab:noisy_density_metric}.
The visual comparisons in Fig.~\ref{fig:results_noisy} also show that our method reconstructs more accurate and smooth surfaces for the noisy point clouds.
More results can be found in Supp.

\subsection{Ablation study}
\label{ablation}
\begin{table}[tb]

\begin{center}
\small
\setlength{\tabcolsep}{0.9mm}{
\begin{tabular}{@{}ll|ll|ccc@{}}
\toprule
DS & SS & LRS & PE & F-score$_\uparrow$ & CD-$L_1$ $_\downarrow$  & NC$_\uparrow$ \\ 
\midrule
\bluecmark &  \bluecmark  & \bluecmark    & \bluecmark  & \textbf{0.943} & \textbf{0.520}  & \textbf{0.960} \\
\midrule
\redxmark  & \bluecmark   & \bluecmark    & \bluecmark  & 0.940 (-0.004) & 0.524 (+0.004) & 0.959 (-0.001) \\
\bluecmark & \redxmark    & \bluecmark    & \bluecmark  & 0.881 (-0.062) & 0.626 (+0.106) & 0.952 (-0.008) \\
\bluecmark &  \bluecmark  &  \redxmark    & \bluecmark  & 0.883 (-0.060) & 0.831 (+0.311) & 0.947 (-0.013) \\
\bluecmark &  \bluecmark  & \bluecmark    & \redxmark   & 0.877 (-0.066) & 0.808 (+0.288) & 0.912 (-0.048) \\ 
\bottomrule
\end{tabular}
}
\end{center}

\vspace{-3mm}
\caption{Ablation study conducted on the Thingi10K~\cite{zhou2016thingi10k} dataset.
SS, DS, LRS, PE represent ``Signed supervision'', ``Derivative supervision'', ``Loss-based importance sampling'', and ``Positional encoding'', respectively.
}
\label{Tab:all_ablation}
\end{table}
\begin{figure}[tb]
\begin{center}
\begin{overpic}[width=1.0\linewidth]{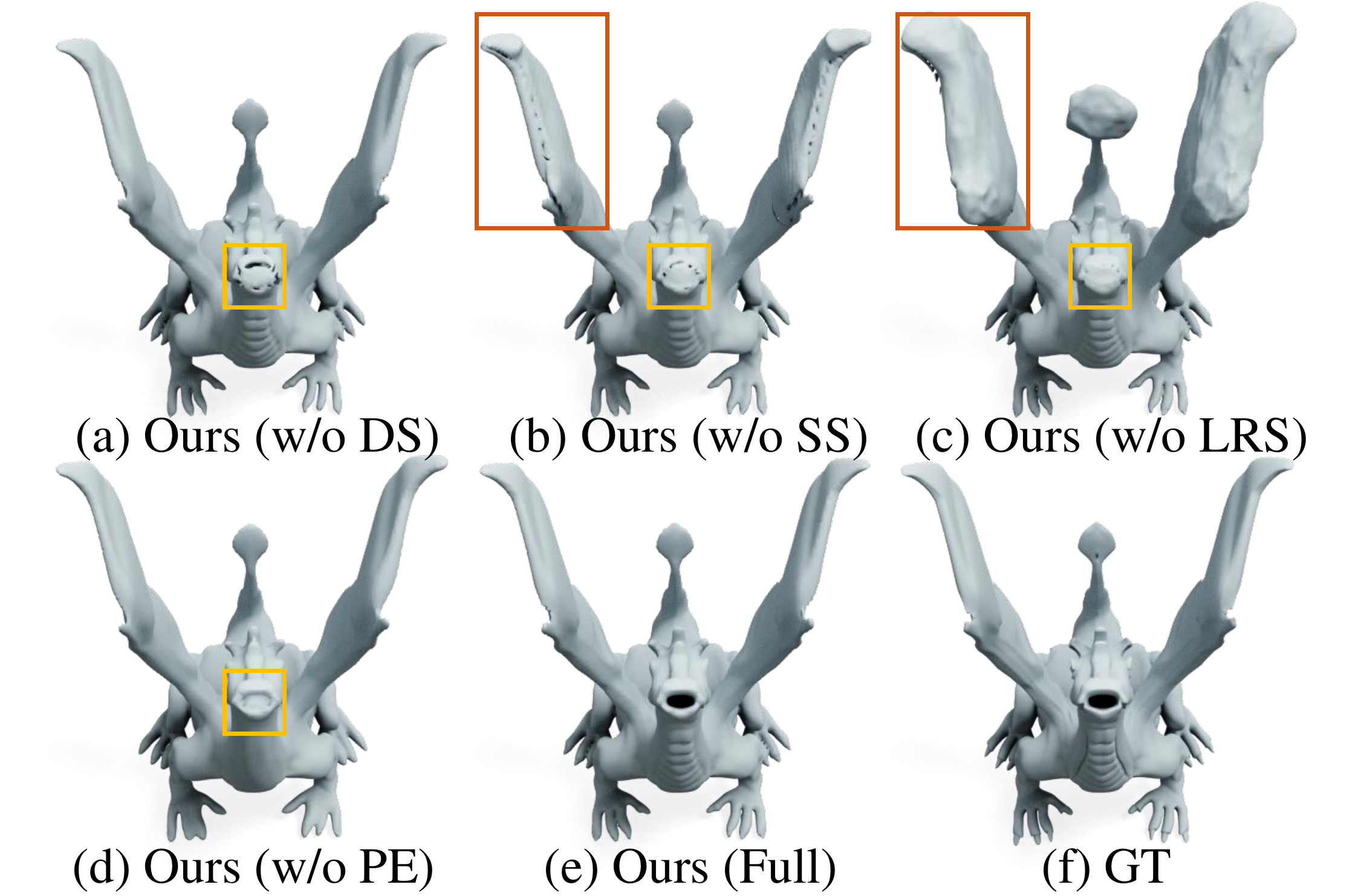}
\end{overpic}
\end{center}
\vspace{-2mm}
\caption
{
Visual ablation. More examples are shown in Supp.
}

\label{fig:visualtization-ablation}
\end{figure}

\paragraph{Signed-supervision \& derivative supervision.}
We conduct experiments to show the effectiveness of signed-supervision (SS) and derivative supervision (DS) in Tab.~\ref{Tab:all_ablation} (1st \& 2nd rows) and Fig.~\ref{fig:visualtization-ablation}. 
For the ablation on SS, we compare the result optimized by our proposed SS with the result by the existing unsigned supervision on the full space.
As expected, removing either SS or DS leads to inaccurate fitting results (in Fig.~\ref{fig:visualtization-ablation} (a) \& (b)).
Specifically, removing the SS introduces more fitting errors in the challenging region (see the ``wing'' part) even with the help of LRS.
\vspace{-5mm}
\paragraph{Loss-based per-region sampling \& positional encoding.}
We conduct ablation to analyze loss-based per-region sampling (LRS) and positional encoding (PE).
For LRS, we compare the result optimized by our LRS with the result optimized by the traditional Gauss sampling (GS) strategy, which is widely used in the existing methods~\cite{gropp2020reg,atzmon2020sal,atzmon2020sald,ma2021pull}.
As shown in Tab.~\ref{Tab:all_ablation} (3rd) and Fig.~\ref{fig:visualtization-ablation} (c), removing LRS in our method will result in significant performance reduction.
Since the sampled points by GS are mostly near to the surface,  this may lead to inefficient sampling in the outside region for adding signed supervision.
Besides, removing PE decreases the overall performance as shown in Tab.~\ref{Tab:all_ablation} (4th) and Fig.~\ref{fig:visualtization-ablation} (d).
\begin{figure}[tb]
\begin{center}
\begin{overpic}[width=1.0\linewidth]{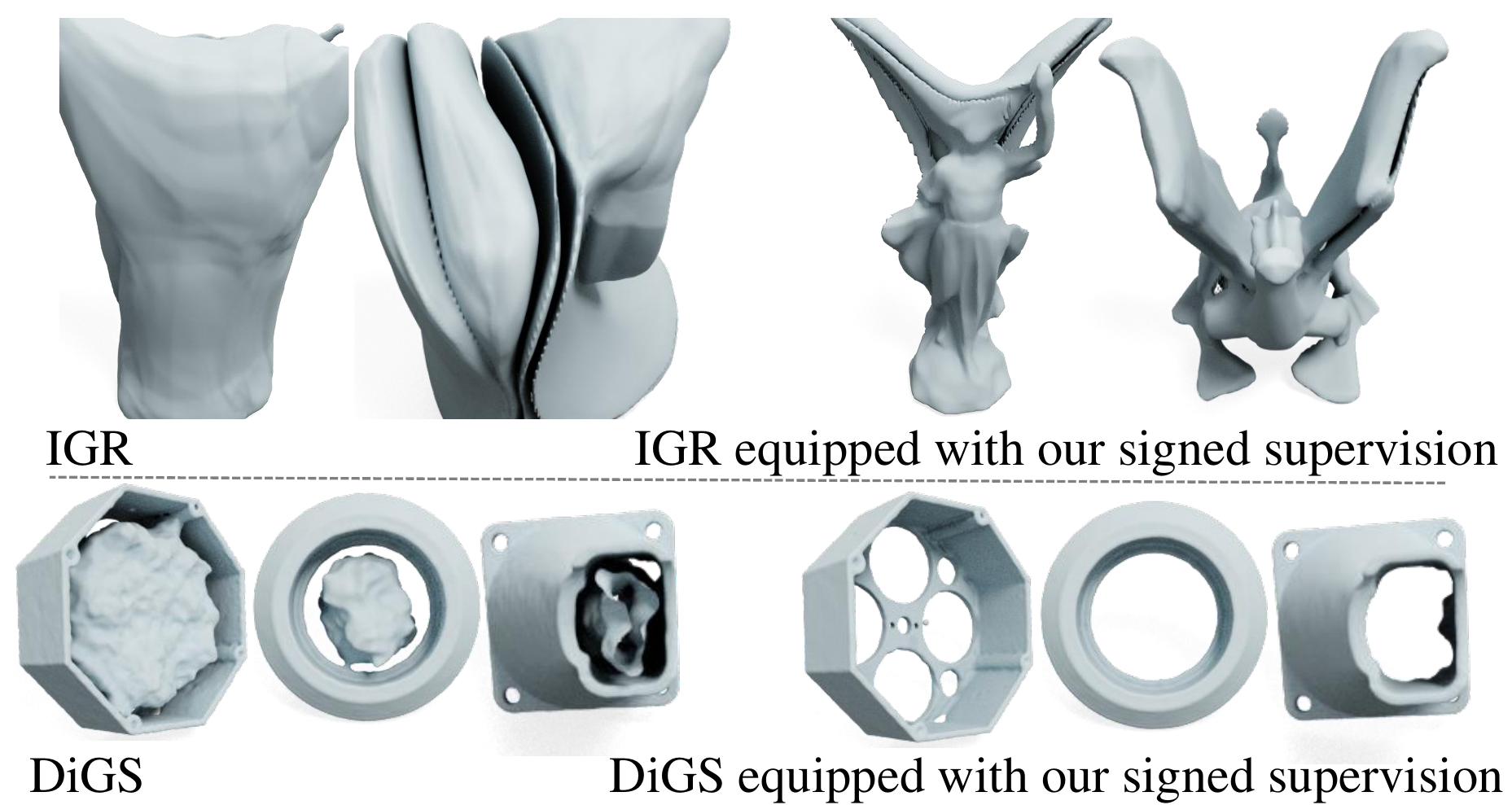}
\end{overpic}
\end{center}
\vspace{-1mm}
\caption{Visual comparisons of results optimized by the original baselines (left) and
the same baselines equipped with our signed supervision (right).
}
\label{fig:signe_supervision}
\end{figure}
\subsection{Further analysis}
\paragraph{Is the proposed signed supervision applicable to other SOTA methods?}
To answer this question, we verify the effectiveness of our signed supervision on two baselines, IGR~\cite{gropp2020reg} and DiGS~\cite{ben2022digs}.
For IGR, we conduct experiments on Thingi10k~\cite{zhou2016thingi10k}, since we observe that IGR had difficulty in fitting coarse shapes for several objects in Thingi10k.
For DiGS, we choose the top 10 hard cases in the ABC subset~\cite{erler2020points2surf} based on the CD-$L_1$ values
obtained experimentally. 
We compare the results optimized with an extra term, \ie, our signed supervision, to the results from the original baselines (IGR and DiGS). 
As shown in Tab.~\ref{Tab:Boosting_IGR} and Tab.~\ref{Tab:Boosting_DiGS}, our signed supervision can significantly improve the performance of both baseline methods (especially on the CD-$L_1$ metric).
Visual comparisons are provided in Fig.~\ref{fig:signe_supervision}.
\begin{table}[tb]
\begin{minipage}{1.0\linewidth}
\centering
\small
\setlength{\tabcolsep}{1.9mm}{
\begin{tabular}{l|cccc}
\toprule
Method  &  Signed Super. &  F-score$_\uparrow$  & CD-$L_1$ $_\downarrow$   & NC$_\uparrow$ \\ 
\hline
IGR~\cite{gropp2020reg} &  & 0.308 & 6.471 & 0.631\\
\hline
IGR~\cite{gropp2020reg} & \bluecmark      & 0.636 & 1.676 & 0.729 \\
\hline
\end{tabular}}
\vspace{1mm}
\caption{Quantitative comparisons of results optimized by IGR and IGR equipped with our signed supervision on Thingi10k~\cite{zhou2016thingi10k}. 
}
\vspace{1mm}
\label{Tab:Boosting_IGR}
\end{minipage}

\begin{minipage}{1.0\linewidth}
\centering
\small
\setlength{\tabcolsep}{1.9mm}{
\begin{tabular}{l|cccc}
\toprule
Method  &  Signed Super. &  F-score$_\uparrow$  & CD-$L_1$ $_\downarrow$   & NC$_\uparrow$ \\ 
\hline
DiGS~\cite{ben2022digs} &  & 0.411 & 4.018 & 0.881\\
\hline
DiGS~\cite{ben2022digs} & \bluecmark      & 0.500 & 2.218 & 0.920  \\
\hline
\end{tabular}}
\vspace{1mm}
\caption{Quantitative comparison of results optimized by DiGS and DiGS equipped with our signed supervision on the ABC~\cite{koch2019abc} hard case subset.
}
\label{Tab:Boosting_DiGS}
\end{minipage}
\end{table}

\vspace{-5mm}
\paragraph{Difficulty of optimization from unoriented point clouds.}
\begin{figure}[tb]
\begin{center}
\begin{overpic}[width=1.0\linewidth]{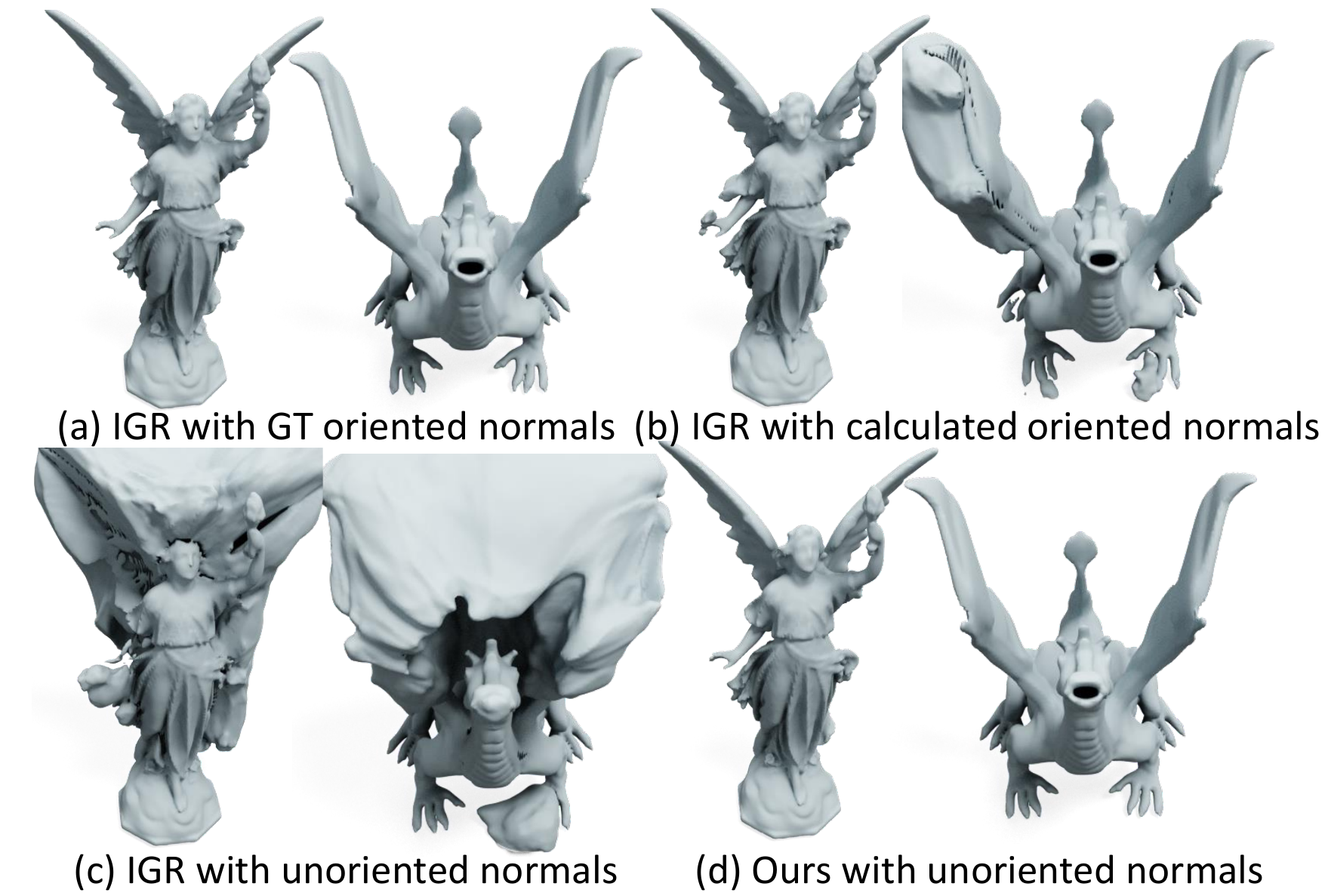}
\end{overpic}
\end{center}
\vspace{-2mm}
\caption{Visual comparisons of results optimized by IGR from different inputs with GT oriented normals, calculated oriented normals and unoriented normals, and ours on Thingi10k~\cite{zhou2016thingi10k}.
}
\label{fig:difficulty_unsigned}
\end{figure}
We conduct experiments to explore the difficulty of optimization from unoriented point clouds in Tab.~\ref{Tab:compare_oriented_normal} and Fig.~\ref{fig:difficulty_unsigned}.
IGR~\cite{gropp2020reg} reconstructs relatively accurate shapes, if GT 
oriented normals are given (1st row in Tab.~\ref{Tab:compare_oriented_normal}).
However, the accurate oriented normals are usually difficult to obtain.
When there is no normal supervision,
the performance decreases significantly (2nd row in Tab.~\ref{Tab:compare_oriented_normal}).
One solution to this problem is that we can adopt traditional methods (\eg, the minimum spanning tree~\cite{zhou2018open3d}) to estimate the normals' orientations.
However, the normal orientation task is notoriously difficult in regions with complicated structures, and the inaccurate normal orientations limit the subsequent reconstruction quality (3rd row in Tab.~\ref{Tab:compare_oriented_normal}).
Another possibility is that we only use the unsigned derivative supervision based on unoriented normals for the optimization.
However, optimization from unsigned supervision occasionally suffers from the difficulty in finding coarse shapes, leading to large fitting errors for complex shapes, as shown in tab.~\ref{Tab:compare_oriented_normal} and Fig.~\ref{fig:difficulty_unsigned}.
In contrast, our SSP achieves a significant improvement with solely unoriented normal information.
\begin{table}[tb]
\begin{center}
\small
\setlength{\tabcolsep}{1.7mm}{
\begin{tabular}{c|l|ccc}
\toprule
Method & Norm. Info.  & F-score$_\uparrow$  & CD-$L_1$ $_\downarrow$   & NC$_\uparrow$  \\ \midrule
IGR & \underline{\textbf{O}}riented (GT)     & 0.917 & 0.785 & 0.94 \\
\midrule
IGR & None & 0.308 & 6.471 & 0.631  \\ 
IGR & \underline{\textbf{O}}riented (Calc.)     & 0.748 & 2.889 & 0.933 \\
\midrule
IGR & \underline{\textbf{U}}noriented    & 0.584 & 4.353 & 0.858 \\  
Ours & \underline{\textbf{U}}noriented    & \textbf{0.943} & \textbf{0.520} & \textbf{0.960} \\
\bottomrule
\end{tabular}
}
\end{center}
\vspace{-2mm}
\caption{Quantitative comparison of results optimized from orientated and unoriented point clouds. "Calc." means "calculated".
}
\label{Tab:compare_oriented_normal}

\end{table}

\vspace{1mm}
\section{Conclusion and Limitation}
We presented SSP for neural surface reconstruction from unoriented point clouds.
First, we propose to utilize the extra coarse signed supervision to help the neural network quickly learn coarse shapes for target (complicated) objects.
To obtain more details, we propose a loss-based region importance sampling strategy and progressive PE to prioritize the optimization.
Our SSP method has achieved improved performance on multiple datasets.
One limitation of our method is it can not handle point clouds with large missing parts (i.e. partial scans).
The reason is two-fold: (i) the calculation of the ``outside'' region is highly unreliable; (ii) the unsigned supervision calculated from partial input is also unreliable.
More results on a dataset (\ie, SRB~\cite{williams2019deep}) containing partial scans are provided in Supp.
%
\clearpage

{\small
\bibliographystyle{ieee_fullname}
\bibliography{egbib}
}

\clearpage
\appendix

\twocolumn[\section*{\LARGE \centering Supplementary Material for \\Semi-signed prioritized neural fitting for surface reconstruction\\ from unoriented point clouds}]


\IncMargin{1em}
\begin{algorithm*}[h!] 
    \caption{Space Partitioning (Python style)}
    \label{algo:space_partition}
    \SetKwInOut{Input}{Input}
    \SetKwInOut{Output}{Output}
    \SetKwData{N}{N}
    \SetKwData{Vocc}{$V_\mathrm{occ}$}
    \SetKwData{Voutside}{$V_\mathrm{outside}$}
    \SetKwData{Vv}{$V_v$}
    \SetKwData{List}{list}
    \SetKwData{neighborPos}{neighbor\_pos}

    \SetKwData{Outside}{outside}
    \SetKwData{Curpos}{cur\_pos}
    \SetKwFunction{checkValid}{check\_valid}
    \SetKwData{Offset}{offset}
    \SetKwData{Index}{index}
    \SetKwComment{Comment}{// }

	\Input{3D array \Vocc, indicating if a voxel is occupied} 
	\Output{3D array \Voutside, indicating if a voxel is outside the object for sure}
	\tcp{shape: $N \times N \times N$}
	\BlankLine 
    Initialize a visited list \Vv\ with 0;
    
    \Outside = \{\}\;
    \BlankLine 
    \tcp{Step 1: find all empty voxels on the six faces of the cube and label them as ``outside''}
    \BlankLine
    \BlankLine 
    \tcp{test voxels on two opposite faces of the cube}
    \For{$i$ in $\{1, N\}$ }{
        \For{$j\leftarrow 1$ \KwTo $N$}{
            \For{$k\leftarrow 1$ \KwTo $N$}{ 
                \If{\Vocc$[i,j,k]== 0$ and \Vv$[i,j,k]==0$}{
                    \Voutside$[i,j,k]$= $1$\;
                    \Outside.append([i,j,k])\;
                }
                \Vv$[i,j,k] = 1$\;
            }    
        }    
    }
    \tcp{test the other two sets of opposite faces similarly}
    \hspace{5mm} $\vdots$

    \BlankLine 
    \tcp{Step 2: a BFS-like procedure to recursively find all ``outside'' voxels connected to the initial outside voxels on the six faces}
    \BlankLine

    $\Offset=\{[1,0,0],[-1,0,0],[0,1,0],[0,-1,0],[0,0,1],[0,0,-1]\}$\;
    \While {$\Index $ < \Outside.size()}{
        \Curpos = \Outside[i]\; 
        \For {$i\leftarrow 1$ \KwTo $6$}{
        \tcp{test all six faces}
            \neighborPos = \{\Curpos[1]+ \Offset[i][1],\Curpos[2]+ \Offset[i][2],\Curpos[3]+ \Offset[i][3]\}\;
            \If{\checkValid (\neighborPos) and $\Vocc[\neighborPos]== 0$ and $\Vv[\neighborPos]==0$}{
                \BlankLine 
                \tcp{if this is a valid \&\& occupied \&\& not visited position}
                \BlankLine 
                $\Voutside[\neighborPos]$= $1$\;
                \tcp{add the connected outside voxel}
                \Outside.append(\neighborPos)\;
            }
            $\Vv[\neighborPos] = 1$\;
            $\Index = \Index + 1$
        }
    }
    \Return \Voutside
 	\end{algorithm*}
 \DecMargin{1em} 
\vspace{1.0in}
In this supplementary material, we provide additional method details (Sec.~\ref{supp:sec:method}), experimental details (Sec.~\ref{supp:sec:exp}), and more results (Sec.~\ref{supp:sec:results}) that could not be fitted into the main paper due of lack of space.

\section{Method details} 
\label{supp:sec:method}

\subsection{Space partitioning}
\label{supp:sec:spacepartition}

We use the same rule to determine the voxel size for all the shapes (ABC subset~\cite{erler2020points2surf}, Thingi10k~\cite{zhou2016thingi10k}, noisy data~\cite{guerrero2018pcpnet}, density-varying data~\cite{guerrero2018pcpnet}, and Surface Reconstruction Benchmark~\cite{williams2019deep}). 
More concretely, for any given shape (represented in the point cloud format and normalized to fit into [-0.9, 0.9]), 
we first find the "distances" from each point to its 50th closest point. 
Then, we calculate the average "distance" (termed as "density\_indicator") from all distance values since the average distance reflects the sampling density of a given shape. 
We use $10 * \mathrm{round}(1/(1.5*  \mathrm{density\_indicator} * 10))$ as the final voxel grid resolution. 
Note that we make sure the resolution is divisible by 10 for convenience. 
Then we mark a voxel as ``occupied'' if it contains at least one point. 
We provide the pseudo code of the space partitioning algorithm (Sec. 3.2 in the main paper) in Supp. Algo.~\ref{algo:space_partition}.

\subsection{Loss-based per-region sampling}

\paragraph{Region-wise loss tracking.} Different losses (i.e., $\mathcal{L}^\mathrm{on}_{\mathrm{dist}}$, $\mathcal{L}^\mathrm{off}_{\mathrm{dist}}$, $\mathcal{L}^\mathrm{on}_{\mathrm{grad}}$, $\mathcal{L}^\mathrm{off}_{\mathrm{grad}}$, $\mathcal{L}_{\mathrm{signed}}$) are applicable to different voxels (Sec. 3.3 in the main paper).
Specifically, we track $\mathcal{L}^\mathrm{on}_{\mathrm{dist}}$ and $\mathcal{L}^\mathrm{on}_{\mathrm{grad}}$ for the 
voxels that contain points.
We track $\mathcal{L}^\mathrm{off}_{\mathrm{dist}}$ and $\mathcal{L}^\mathrm{off}_{\mathrm{grad}}$ for all voxels in region $\mathcal{V}_{\mathrm{uncertain}}$ .
We track $\mathcal{L}_{\mathrm{signed}}$ for all voxels in region $\mathcal{V}_{\mathrm{known}}$ .

\paragraph{Adaptive sampling.} 
As mentioned in the main paper (Sec 3.3), we perform a two-step sampling for each loss: (i) adaptively sample a set of voxels
based on the previous region losses and (ii) sample points
within each region (voxels).
Specifically, for the sampling of on-surface points and off-surface points, we additionally randomly sample voxels to  better cover the whole $V_\mathrm{uncertain}$ region.
Besides, we keep a similar number of sample points (34588) as IGR~\cite{gropp2020reg} (34816).
Specifically, we set $16384 * \frac{7}{9}$ for the sample number of on-surface points, $16384 * \frac{1}{3}$ as the sample number of off-surface points in region $V_\mathrm{uncertain}$ and $16384$ as the sample number in the outside region $V_\mathrm{outside}$.

\subsection{Motivation for using derivative supervision in the free space}

According to the following proposition, the normal of point $q$'s nearest point $p'$ (denoted as $\frac{\nabla f_\theta(p')}{||\nabla f_\theta(p')||}$) on the underlying continuous surface equals to the derivative of $q$. 
Empirically, we found it helpful
to utilize the normal of the nearest point $p$ (denoted as $n_p$) in $P$ to approximate $\frac{\nabla f_\theta(p')}{||\nabla f_\theta(p')||}$ as a supervision for point $q$ in the free space.
\begin{wrapfigure}[6]{r}{0.25\columnwidth}
\vspace{5pt}
\hspace{-15pt}
\includegraphics[width=0.28\columnwidth]{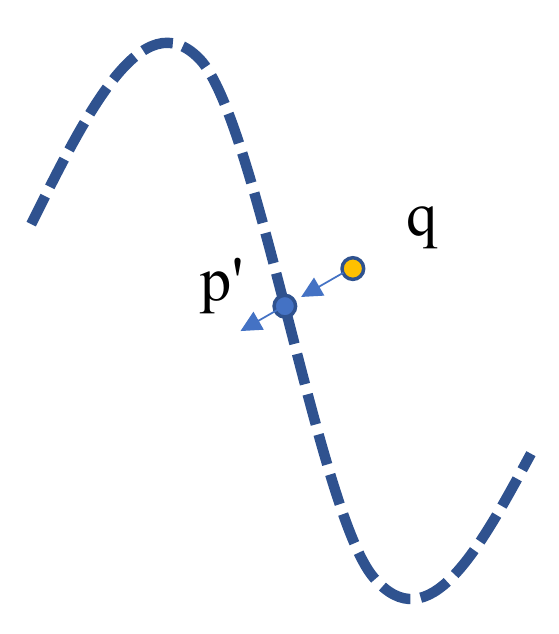}
\vspace{0pt}
\label{fig:wl_example}
\end{wrapfigure} 
%

\begin{prop}
    Given a random point $q$ in the free space and its closest point on the underlying surface $p'$,
    we have $\frac{\nabla f(p')}{||\nabla f(p')||} = \frac{\nabla f(q)}{||\nabla f(q)||}$, where the derivative always exists for $f$.
    \label{prop}
\end{prop}

\noindent\textbf{The proof:}
we first consider point $q$ that is inside the underlying shape, which means $f(q)<0$.

(1) First, we prove the following proposition: 
\begin{equation}
\frac{\nabla f(p')}{\left\|\nabla f(p')\right\|}=\frac{q - p'}{\|q - p'\|},
\end{equation}
where $\|\cdot\|$ is L2 norm. 
Since $p'$ is $q$'s closest point on the underlying surface defined by $f(x)=0$, then
\begin{equation*}
p' = \underset{ \mathrm{x}}{\operatorname{argmin}}\|x-q\| \;\;s.t.\;\; f(x)=0.
\end{equation*}
And the corresponding Lagrange function is 
\begin{equation*}
L(x)=\|x-q\|+\lambda f(x),
\end{equation*}
whose gradient can be calculated as
\begin{equation*}
\nabla_{(p', \lambda)} L=\left(\frac{-(q-p')}{\|q- p' \|}+\lambda \nabla f(p'), f(p')\right).
\end{equation*}
Let $\frac{-(q-p')}{\|q-p'\|}+\lambda \nabla f(p')=0$ 
and we get $\nabla f(p')=\lambda \frac{q-p'}{|| q-p' \|}$.
Hence, we know that $\nabla f(p')$ is collinear to vector $(q-p')$.
Since the gradient denotes the direction of the greatest increase of the function, we have $\frac{\nabla f(p')}{\left\|\nabla f(p')\right\|}=\frac{q-p'}{\|q-p'\|}$.

(2) Then, we can prove that
\begin{equation}
\frac{\nabla f(q)}{\left\|\nabla f(q)\right\|}=\frac{q-p'}{\|q-p'\|} = \frac{\nabla f(p')}{\left\|\nabla f(p')\right\|}.
\end{equation}

According to the definition of directional derivative, 
\begin{equation*}
\nabla_{v} f(x)=\lim _{h \rightarrow 0} \frac{f(x+h v)-f(x)}{h},
\end{equation*}
where $v$ is the given unit vector. 
The relation between gradient and directional derivative at point $q$ can be written as 
\begin{equation*}
\frac{\nabla f(q)}{\left\|\nabla f(q)\right\|}=  \underset{ v}{\operatorname{argmax}} \nabla_{v} f(q)
\end{equation*}
and
\begin{equation*}
\left\|\nabla f(q)\right\|=  \underset{ v}{\operatorname{max}} \nabla_{v} f(q).
\end{equation*}
%
Since the signed distance function should satisfy $\|\nabla f(q)\|=1$ 
, we can know that the directional derivative always satisfies $\nabla_{v} f(q) \leq \|\nabla f(q)\| = 1$ for any direction $v$. 
Moreover, by definition, $\nabla_{v} f(q)$ can be expressed as 
\begin{equation*}
\nabla_{v} f(q)=\lim _{h \rightarrow 0} \frac{f(q+h v)-f(q)}{h}.
\end{equation*}

We can consider the specific direction $v'=\frac{q-p'}{\|q-p'\|}$ and let $q^{*}= q+hv'$. 
We also denote 
\begin{equation*}
p''= \underset{p s.t. f(p)=0}{\operatorname{argmin}}\|q^*-p\|. 
\end{equation*}
We can prove that $p''=p'$ by contradiction, i.e., if $p'' \neq p'$, then $\|q^{*}-p''\|$ < $\|q^{*}-p'\|$, we add the same value on each side of the inequality, $\|q^{*}-q\| + \|q^{*}-p''\| < \|q^{*}-q\| + \|q^{*}-p'\| = \|q-p'\|$, since $q^*$ locates in the line segment $(q,p')$. 
However, $\|q^{*}-q\| + \|q^{*}-p''\| \geq \|q-p''\| > \|q-p'\|$ which leads to contradiction. Therefore,  $p''=p'$.

Thus, 
\begin{equation*}
\begin{split}
\nabla_{v'} f(q) & =\lim _{h \rightarrow 0} \frac{f(q^{*})-f(q)}{h} \\ & = \lim _{h \rightarrow 0} \frac{-1 *\|q^{*}-p'\|-(-1) *\|q-p’\|)}{h} \\ &= 1.
\end{split}
\end{equation*}
Hence, the direction $v'$ yields $\underset{v}{\operatorname{argmax}} \nabla_{v} f(q)$. Therefore, we obtain
\begin{equation}
    \frac{\nabla f(q)}{\left\|\nabla f(q)\right\|}=\frac{q-p'}{\|q-p'\|}.
\end{equation}

In the meantime, if $q$ is outside the shape, it can be proved in a similar way that
\begin{equation*}
\frac{\nabla f(p)}{\left\|\nabla f(p)\right\|}=\frac{\nabla f(q)}{\left\|\nabla f(q)\right\|} = \frac{- (q-p')}{\|q-p'\|}.
\end{equation*}

\section{Experimental details}
\label{supp:sec:exp}

\paragraph{Implementation details of the proposed semi-signed prioritized (SSP) neural fitting.}
We adopt the MLP architecture that contains eight hidden layers, each with 512 hidden units, to represent the SDF function as IGR~\cite{gropp2020reg}.
Besides, we adopt Geometry Network Initialization (GNI)~\cite{atzmon2020sal} to initialize the implicit SDF function to a rough unit sphere. 
We use ADAM optimizer~\cite{kingma2015adam} with an initial learning rate of 0.005 and schedule the learning rate to decrease by a factor of 0.5 every 2000 epochs.
For the whole experiment except for the noisy data (\ie, ABC subset~\cite{erler2020points2surf}, Thingi10K~\cite{zhou2016thingi10k}, density-varying data~\cite{guerrero2018pcpnet}, noisy data~\cite{guerrero2018pcpnet}, and Surface Reconstruction Benchmark~\cite{williams2019deep}) , the weight parameter ${w_i}$ is $\{40,20,1,1,1,10\}$ in Eq.~(8) in the main paper. 
For the noisy data, the weight parameter is adjusted to $\{20,10,20,10,1,10\}$ to rely less on the distance losses ($\mathcal{L}^\mathrm{on}_{\mathrm{dist}}, \mathcal{L}^\mathrm{free}_{\mathrm{dist}}$)
and more on the gradient losses ($\mathcal{L}^\mathrm{on}_{\mathrm{grad}}, \mathcal{L}^\mathrm{free}_{\mathrm{grad}}$).

\paragraph{Optimization setting.}
We list the optimization epochs and fitting time for different methods in Supp.~Tab.~1.
We optimize our method on each shape for $10K$ epochs.
For DiGS~\cite{gropp2020reg}, we optimize the neural network using the default maximum number of epoch, \ie, $10K$.
For Neural-Pull~\cite{ma2021pull}, we optimize the neural network using the default maximum number of epoch $40K$ as in their official code.
We optimize IGR~\cite{gropp2020reg}, SAL~\cite{atzmon2020sal}, and SALD~\cite{atzmon2020sald} for $20K$ epoch to keep a similar optimization time as our method.
Since SALD~\cite{atzmon2020sald} did not release the code for surface reconstruction, we implement the code based on SAL~\cite{atzmon2020sal} by adding the on-surface derivative supervision~\cite{atzmon2020sald} with the default loss weight according to the SALD~\cite{atzmon2020sald} paper.
For SAP~\cite{peng2021shape}, we just follow the provided optimization-based setting to report the results. 
\begin{table}[h!]

\begin{center}
\small
\setlength{\tabcolsep}{0.8mm}{
\begin{tabular}{l|cccccc}
\toprule
Method & Ours & DiGS~\cite{ben2022digs}   & IGR~\cite{gropp2020reg}    & SAL~\cite{atzmon2020sal}   & SALD~\cite{atzmon2020sald} & NP~\cite{ma2021pull}   \\
\midrule
Epoch & 10K & 10K &  20K & 20K  & 20K & 40K  \\ 
Time  & $\thicksim$15 & $\thicksim$15 & $\thicksim$27 & $\thicksim$20 & $\thicksim$22 & $\thicksim$15  \\
\bottomrule
\end{tabular}}
\label{tab:optimization_epoch}
\end{center}
\caption{The optimization epochs \& approximate time (in minutes) for different optimization-based neural methods.
}
\end{table}

\paragraph{Evaluation metrics.} 
We follow the same procedure as in SAP~\cite{peng2021shape} to calculate the metrics, including the L1-based Chamfer Distance (denoted as CD-$L_{1}$), the Normal Consistency (NC), and the F1-score with default threshold $1\%$ in Sec.~4.1 of the main paper.
Specifically, for the ABC subset~\cite{erler2020points2surf}, we randomly sample $40,000$ points, each on the reconstructed mesh and the ground-truth mesh, respectively, to calculate the aforementioned metrics.
For Thingi10K~\cite{zhou2016thingi10k}, we randomly sample $50,000$ points, each on the reconstructed mesh and the ground-truth mesh, respectively, to calculate the metrics, following SAP~\cite{peng2021shape}.
For the density-varying data and noisy data, we directly use the provided clean point cloud points (containing $100,000$ points) 
as GT and randomly sample the same number of points on the reconstructed meshes to calculate the metrics.

\paragraph{Details for applying proposed signed supervision to existing methods.} 
To apply our proposed supervision, we use the space partition algorithm in Supp.~Sec.~\ref{supp:sec:spacepartition} to find the $V_\mathrm{known}$ region.
Then, we randomly sample points in the $V_\mathrm{known}$ region to apply the signed guidance in addition to the original constraints of the official methods (\ie, IGR~\cite{gropp2020reg}, DiGS~\cite{ben2022digs}) without any other change.
Specifically, we set the number of the sample points in the $V_\mathrm{known}$ region as 16348 for IGR~\cite{gropp2020reg} and DiGS~\cite{ben2022digs}.

\section{More results}
\label{supp:sec:results}

\begin{figure*}[tbh]
\begin{center}
\begin{overpic}[width=1.0\linewidth]{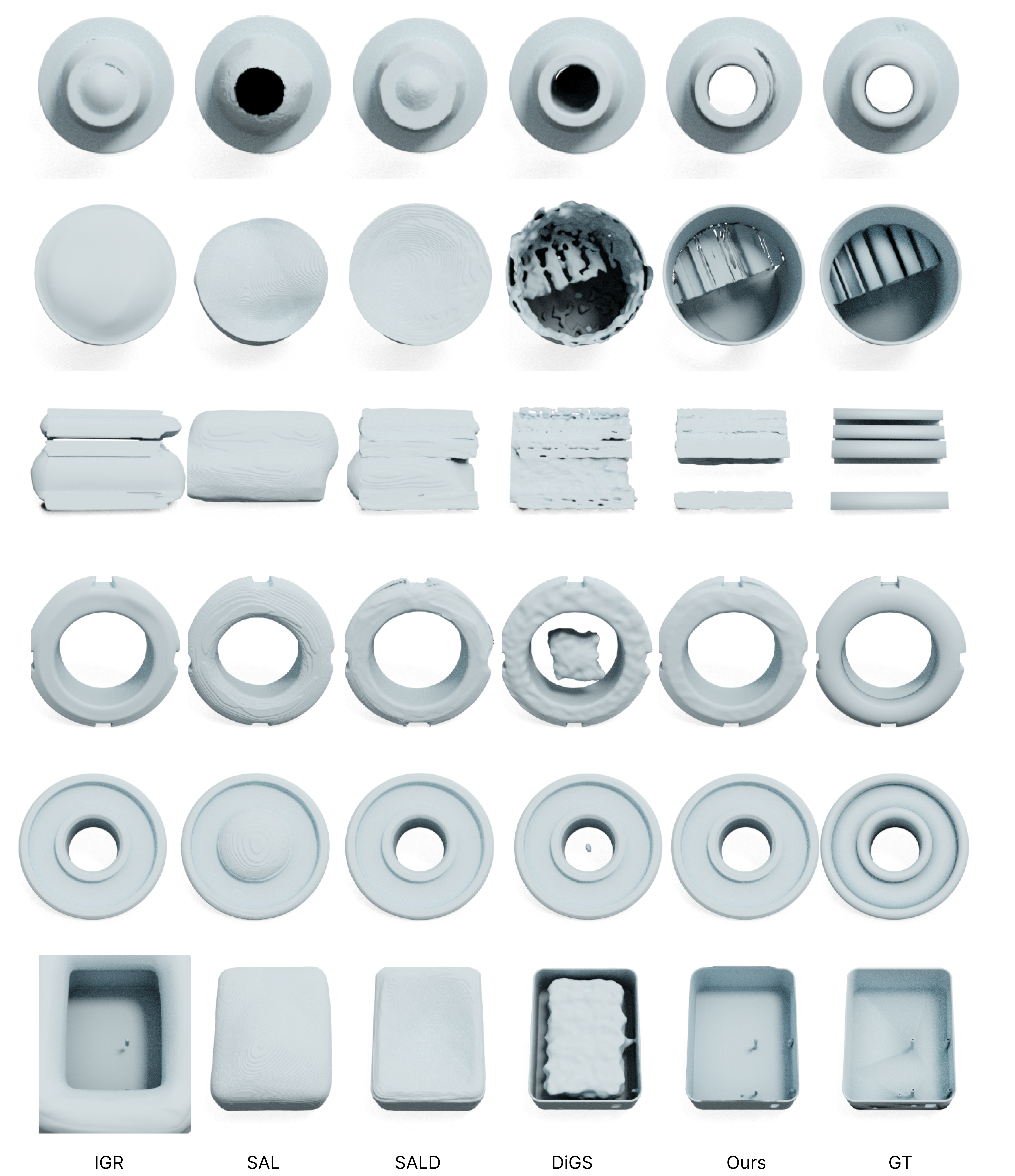}
\end{overpic}
\end{center}
\vspace{-1mm}
\caption{Visual comparisons for neural methods (IGR~\cite{gropp2020reg}, SAL~\cite{atzmon2020sal}, SALD~\cite{atzmon2020sald}, and DiGS~\cite{ben2022digs}) on the ABC subset~\cite{erler2020points2surf}.
}
\label{fig:ghost-surface}
\end{figure*}

\begin{figure*}[tbh]
\begin{center}
\begin{overpic}[width=1.0\linewidth]{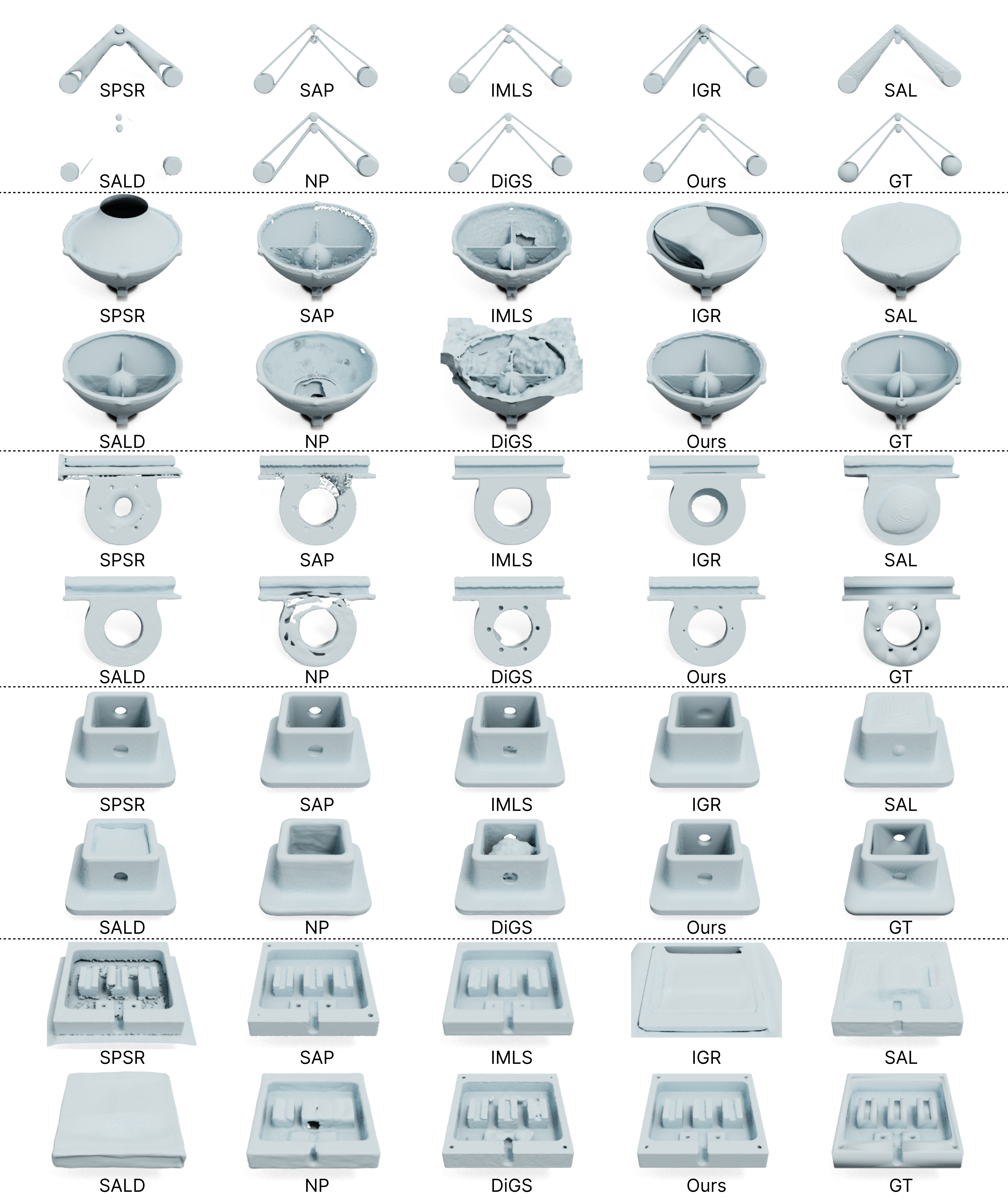}
\end{overpic}
\end{center}
\vspace{-2mm}
\caption{Visual comparisons with more methods on the ABC subset~\cite{erler2020points2surf}.
}
\label{fig:more_abc_data}
\end{figure*}

\begin{figure*}[tbh]
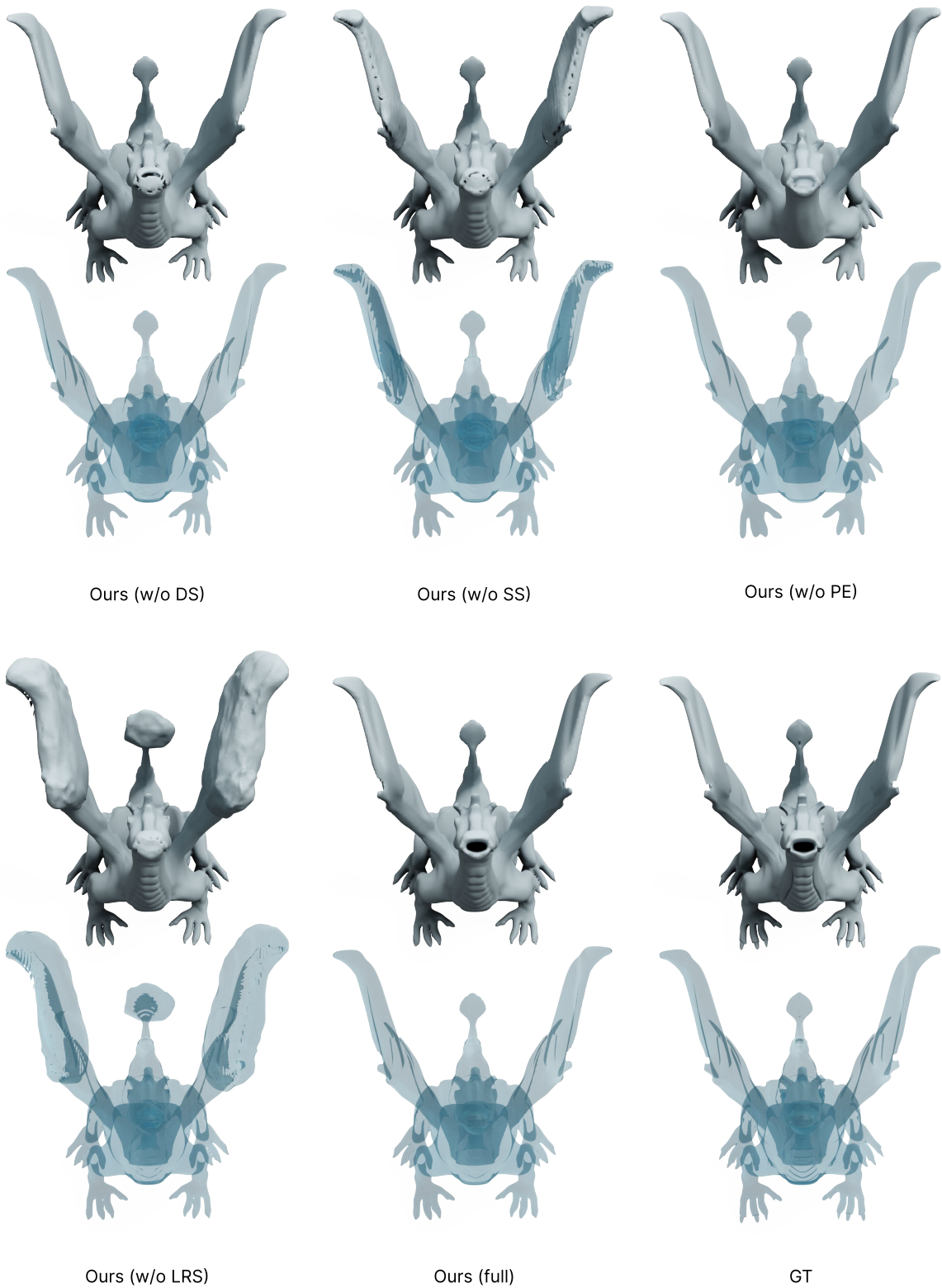

\begin{center}
\begin{overpic}[width=0.95\linewidth]{./supp_image_small/Ablation_study_old.pdf}
\end{overpic}
\end{center}
\vspace{-7mm}
\caption{Ablation study on Thingi10K~\cite{erler2020points2surf}. Note that, we provide two images (\ie, transparent image (top) and non-transparent image (bottom)) for each result.
}
\label{fig:ablation_study_old}
\end{figure*}

\paragraph{Main examples.}
As shown in Supp. Fig.~\ref{fig:ghost-surface}, we provide more visual comparisons (Sec 3.2 and Sec 3.3 in the main paper), 
including IGR~\cite{gropp2020reg}, SAL~\cite{atzmon2020sal}, SALD~\cite{atzmon2020sald}, DiGS~\cite{ben2022digs}, and ours.
The results demonstrate that our method is able to reconstruct more accurate surfaces while producing less artifacts.
Yet, existing neural methods~\cite{gropp2020reg, atzmon2020sal, atzmon2020sald,ben2022digs} occasionally generate ghost surfaces.
In addition, we provide more comparison examples on Thingi10K~\cite{zhou2016thingi10k}, the density-variation data~\cite{guerrero2018pcpnet}, and the noisy data~\cite{guerrero2018pcpnet} in Supp.~\cref{fig:thingi10k_srb_data,fig:density_data,fig:noisy_data}, respectively.
Besides, we provide visual comparisons with more baseline methods on the examples shown in the main paper; please see Supp.~\cref{fig:more_abc_data,fig:thingi10k_srb_data_old_1,fig:thingi10k_srb_data_old_2,fig:density_data_old,fig:noisy_data_old}.
\paragraph{Ablation experiment.}
We provide more visual comparisons for the ablation study in Supp. ~\cref{fig:ablation_study,fig:ablation_study_old},
in which the transparent visualization results show that removing the signed supervision wrosens the accuracy on thin structures, thereby leading to undesired surfaces.

\paragraph{Results on Surface Reconstruction Benchmark ~\cite{williams2019deep} data.}
\begin{table}[tb]
\begin{center}
\small
\setlength{\tabcolsep}{1.8em}{
\begin{tabular}{lcc}
\toprule
 & \multicolumn{2}{c}{SRB~\cite{williams2019deep}} \\
\cmidrule(lr){2-3} 
Methods &    F-score $\uparrow$ &  CD-$L_1$ $\downarrow$  \\ 
\hline
SPSR~\cite{kazhdan2013screened} &   0.779 & 0.871 \\
SAP~\cite{peng2021shape}     &  0.781&0.852  \\\hline
IMLS~\cite{liu2021deep} & 0.714 & 0.933 \\\hline
N-P~\cite{ma2021pull}      &  0.582 & 1.204   \\
SAL~\cite{atzmon2020sal} & 0.722 & 1.194 \\
SALD~\cite{atzmon2020sald} & 0.518 & 1.812 \\
IGR~\cite{gropp2020reg} & 0.647 & 2.872  \\
DiGS~\cite{ben2022digs}    & \textbf{0.790} & \textbf{0.800} \\
\hline
Ours   & 0.783 & 0.825 \\
\bottomrule
\end{tabular}
}
\end{center}
\vspace{-3mm}
\caption{
Quantitative results on SRB~\cite{williams2019deep}. 
Note that the SRB dataset does not contain GT normal to calculate the normal consistency (NC) metric. 
}
\label{tab:SRB}
\end{table}

We conduct the experiments on the Surface Reconstruction Benchmark (SRB)~\cite{williams2019deep} data.
Note that, we randomly sample $50,000$ points, each on the reconstructed mesh and the ground-truth dense point clouds, respectively, to calculate the metrics, following SAP~\cite{peng2021shape}.
The quantitative results are provided in Supp. Tab.~\ref{tab:SRB} and the visualization results are shown in Supp. Fig.~\ref{fig:srb}.
Overall, SSP achieves comparable performance with the current SOTA methods (DiGS~\cite{ben2022digs}) on SRB~\cite{williams2019deep}. 
We notice that the input point clouds in SRB~\cite{williams2019deep} contain some missing regions, leading to unreliable unsigned supervisions.
In this case, SSP may wrongly prioritize the optimization towards the missing regions according to the tracked loss values, thereby causing large errors.
See the top example in Supp. Fig.~\ref{fig:srb}.
We regard this as a limitation of our method and leave this as a future work to introduce the data prior for accurate surface reconstruction from inputs with missing regions.

\begin{figure*}[tbh]
\begin{center}
\begin{overpic}[width=1.0\linewidth]{./supp_image_small/Chanllenging_data_new.pdf}
\end{overpic}
\end{center}
\vspace{-7mm}
\caption{Visual comparisons on Thingi10K~\cite{erler2020points2surf}.
}
\label{fig:thingi10k_srb_data}
\end{figure*}

\begin{figure*}[!tbh]
\begin{center}
\begin{overpic}[width=0.95\linewidth]{./supp_image_small/density_data_new.pdf}
\end{overpic}
\end{center}
\vspace{-5mm}
\caption{Visual comparisons on the density-variation data~\cite{guerrero2018pcpnet}.
}
\label{fig:density_data}
\end{figure*}

\begin{figure*}[!tbh]
\begin{center}
\begin{overpic}[width=1.0\linewidth]{./supp_image_small/Noisy_new.pdf}
\end{overpic}
\end{center}
\vspace{-3mm}
\caption{Visual comparisons on the noisy data~\cite{guerrero2018pcpnet}.
}
\label{fig:noisy_data}
\end{figure*}

\begin{figure*}[tbh]
\begin{center}
\begin{overpic}[width=0.9\linewidth]{./supp_image_small/Chanllenging_data_old_1.pdf}
\end{overpic}
\end{center}
\vspace{-7mm}
\caption{Visual comparisons on Thingi10K~\cite{erler2020points2surf}.
}
\label{fig:thingi10k_srb_data_old_1}
\end{figure*}

\begin{figure*}[tbh]
\begin{center}
\begin{overpic}[width=1.0\linewidth]{./supp_image_small/Chanllenging_data_old_2.pdf}
\end{overpic}
\end{center}
\vspace{-7mm}
\caption{Visual comparisons on Thingi10K~\cite{erler2020points2surf}.
}
\label{fig:thingi10k_srb_data_old_2}
\end{figure*}

\begin{figure*}[!tbh]
\begin{center}
\begin{overpic}[width=0.95\linewidth]{./supp_image_small/density_data_old.pdf}
\end{overpic}
\end{center}
\vspace{-5mm}
\caption{Visual comparisons on the density-variation data~\cite{guerrero2018pcpnet}. 
}
\label{fig:density_data_old}
\end{figure*}

\begin{figure*}[!tbh]
\begin{center}
\begin{overpic}[width=1.0\linewidth]{./supp_image_small/Noisy_data_old.pdf}
\end{overpic}
\end{center}
\vspace{-3mm}
\caption{Visual comparisons on the noisy data~\cite{guerrero2018pcpnet}. 
}
\label{fig:noisy_data_old}
\end{figure*}

\begin{figure*}[tbh]
\begin{center}
\begin{overpic}[width=0.95\linewidth]{./supp_image_small/Ablation_study.pdf}
\end{overpic}
\end{center}
\vspace{-7mm}
\caption{Ablation study on Thingi10K~\cite{erler2020points2surf}. Note that, we provide two images (\ie, non-transparent image (top) and transparent image (bottom)) for each result.
}
\label{fig:ablation_study}
\end{figure*}

\begin{figure*}[tbh]
\begin{center}
\begin{overpic}[width=0.95\linewidth]{./supp_image_small/SRB.pdf}
\end{overpic}
\end{center}
\vspace{-7mm}
\caption{Visual comparisons on the SRB~\cite{williams2019deep} data. For the GT shape in the top example, the red color encodes the distance between the GT point clouds with the input clouds. Deeper red color means the larger distance, thus indicating the missing regions. 
}
\label{fig:srb}
\end{figure*}

\end{document}